\pdfoutput=1

\documentclass[11pt]{article}

\usepackage[]{acl}
\usepackage{times}
\usepackage{latexsym}
\usepackage{makecell}
\usepackage{bm}
\usepackage[T1]{fontenc}

\usepackage[utf8]{inputenc}
\usepackage{subfigure}
\usepackage{microtype}
\usepackage{enumitem}
\usepackage{graphicx}
\usepackage{multirow}
\usepackage{amsmath}
\usepackage{amssymb}
\usepackage{algorithm}
\usepackage{algorithmicx}
\usepackage{algpseudocode}
\usepackage{xcolor}
\usepackage{array}
\usepackage{booktabs}
\usepackage{arydshln}
\title{Focused Large Language Models are Stable Many-Shot Learners}
\author{Peiwen Yuan$^1$, Shaoxiong Feng$^2$, Yiwei Li$^1$, Xinglin Wang$^1$, Yueqi Zhang$^1$\\ {\bf Chuyi Tan$^1$, Boyuan Pan$^2$, Heda Wang$^2$, Yao Hu$^2$, Kan Li$^{1}$\footnotemark[1]}\\
  $^1$School of Computer Science and Technology, Beijing Institute of Technology \\
  $^2$Xiaohongshu Inc \\
  \texttt{\{peiwenyuan,liyiwei,wangxinglin,zhangyq,tanchuyi,likan\}@bit.edu.cn} \\
  \texttt{\{shaoxiongfeng2023,whd.thu\}@gmail.com} \ \  \texttt{\{panboyuan,xiahou\}@xiaohongshu.com}}

\begin{document}
\maketitle
\renewcommand{\thefootnote}{\fnsymbol{footnote}} 
\footnotetext[1]{Corresponding author.} 
\renewcommand{\thefootnote}{\arabic{footnote}}
\begin{abstract}
In-Context Learning (ICL) enables large language models (LLMs) to achieve rapid task adaptation by learning from demonstrations.
With the increase in available context length of LLMs, recent experiments have shown that the performance of ICL does not necessarily scale well in many-shot (demonstration) settings.
We theoretically and experimentally confirm that the reason lies in more demonstrations dispersing the model attention from the query, hindering its understanding of key content. 
Inspired by how humans learn from examples, we propose a training-free method \textsc{FocusICL}, which conducts triviality filtering to avoid attention being diverted by unimportant contents at token-level and operates hierarchical attention to further ensure sufficient attention towards current query at demonstration-level.
We also design an efficient hyperparameter searching strategy for \textsc{FocusICL} based on model perplexity of demonstrations.
Comprehensive experiments validate that \textsc{FocusICL} achieves an average performance improvement of 5.2\% over vanilla ICL and scales well with many-shot demonstrations.

\end{abstract}

\section{Introduction}
The rapid development of large language models (LLMs) has facilitated the emergence and enhancement of their In-Context Learning (ICL) abilities \citep{emergent,ICLsurvey}. 
As a training-free method, ICL can achieve fast model adaptation on specific tasks based on several demonstrations prefixed to the query, formally denoted as $\texttt{ICL}_{}(response|demos,query)$. 
Intuitively, more demonstrations can help LLMs better understand the task and increase the likelihood of finding demonstrations that aid in responding queries, thus leading to better performance. 
Theoretically, a similar conclusion can be drawn. Previous studies \citep{ICLasFine,ICLasFine2,ICLasFine3,ICLasFine4} have theoretically inferred that ICL can be viewed as an implicit finetuning process, with demonstrations analogous to training samples. 
On this basis, as finetuning has been validated to comply with the scaling law \citep{FineScal} where performance increases with the number of training samples, the performance of ICL should also positively correlates with the number of demonstrations, which has been experimentally verified by previous studies \citep{ICLscale,ICLscale2}. 

\begin{figure}[t]
\centering
\includegraphics[width=0.48\textwidth]{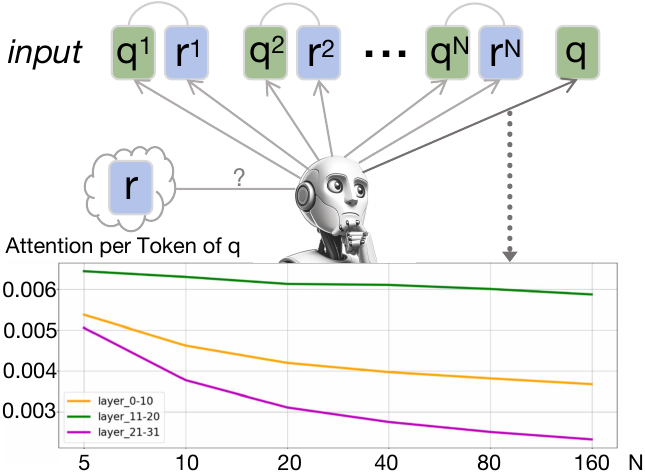} 
\caption{The average model attention for query is dispersed by the increased number of demonstrations, causing inadequate understanding of query.}
\label{fig:abs}
\end{figure}

However, with the increase in available context length of LLMs \citep{gemini15}, some studies \citep{controller,manyICL} observe counterexamples when scaling the demonstration numbers from few-shot to many-shot. \citet{manyICL} finds that the optimal number of demonstrations for six out of eleven benchmarks is not the maximum number they have tested. Our experimental results (Figure~\ref{fig:scale}) also indicate that the model performance might decline with increased demonstrations when applying ICL, exhibiting an inverse-scaling phenomenon \citep{inversescale}. These findings indicate that LLMs are not stable many-shot learners.

To understand this gap, we revisit the derivation of \citet{ICLasFine} that formally equates ICL with finetuning and identify that their approximation of standard attention operation as linear attention operation will ignore the competition for attention between demonstrations and the query when generating the response. Since this approximation is key to the equivalence of ICL and finetuning, we hypothesize that the reason why ICL does not adhere to the scaling law like finetuning is that more demonstrations can divert attention away from the query. 
Inadequate attention and understanding of the query can naturally lead to inferior response. 
To verify our hypothesis, we first conduct experiments confirming that increasing the number of demonstrations does lead to a decrease in model attention towards queries (Figure~\ref{fig:abs}). We further experiment by adding blank spaces within the demonstrations and confirm that: the more blank spaces added, the more attention towards queries distracted by blanks, resulting in lower response accuracy (Figure~\ref{fig:pad}).

Inspired by the way humans benefit from ignoring irrelevant contents and integrating insights from multiple examples when solving problems, we propose \textsc{FocusICL} to avoid the attention dispersion issue faced by ICL. 
Specifically, at the token-level, \textsc{FocusICL} conducts triviality filtering by adaptively masking unimportant tokens of demonstrations based on attention distribution, allocating the attention to more important contents. 
At the demonstration-level, \textsc{FocusICL} performs hierarchical attention mechanism by dividing demonstrations into multiple batches and respectively conducting intra-batch and inter-batch attention operations. 
The limited demonstration number within each batch ensures sufficient attention to the query, while inter-batch attention integrates the benefits from a larger number of demonstrations.
We further introduce an efficient hyperparameter searching strategy for \textsc{FocusICL} according to model perplexity of demonstrations.

Our experiments across three LLMs on five benchmarks confirm that \textsc{FocusICL} achieves an average performance improvement of 5.2\% over ICL by avoiding attention dispersion, with lower inference overhead. This demonstrates the effectiveness, efficiency, and generalizability of \textsc{FocusICL}. 
Furthermore, we observe that \textsc{FocusICL} achieves performance scaling with the number of demonstrations by maintaining attention on critical parts, making demonstration number a possible scaling direction for LLM-based AGI.
Finally, we propose a unified perspective to understand the divergent phenomena observed in previous studies, where more demonstrations lead to either improved \citep{ICLscale} or deteriorated \citep{manyICL} performance in ICL. 
Based on experimental results, we conclude that the performance of ICL initially benefits but subsequently suffers from more demonstrations. 
The weaker the model and the closer the relationship between samples, the later the sweet spot for the number of demonstrations occurs.

Our contributions are summarized as follows:
\begin{enumerate}[itemsep=1mm, parsep=0mm]

\item We analyze that the reason more demonstrations may lead to a decline in ICL performance is that they degrade the model understanding of query by dispersing its attention.

\item We propose \textsc{FocusICL} to achieve rational attention allocation via triviality filtering operation and hierarchical attention mechanism, making LLMs stable many-shot learners.

\item We conduct comprehensive experiments and analyses to validate the effectiveness, efficiency, generalizability and scalability of \textsc{FocusICL}.

\end{enumerate}
\section{Background}
\paragraph{Formalization of ICL}
We follow \citep{ICLsurvey} to define the general ICL paradigm.
Given an LLM $\boldsymbol{\mathcal{M}}$ and a query $\boldsymbol{q}$, we choose $N$ demonstrations from a candidate set $\boldsymbol{\mathcal{S}_{demos}}=\{(\boldsymbol{q_i},\boldsymbol{r_i})\}_{i=1}^{M}$ to attain the response $\boldsymbol{r}$ from $\boldsymbol{\mathcal{M}}$ as follows:
\begin{equation}
    \small
    \boldsymbol{r} = \texttt{Sampling}(\boldsymbol{\mathcal{M}}(\texttt{Cat}[\underbrace{\boldsymbol{q_0};\boldsymbol{r_0};...;\boldsymbol{q_N};\boldsymbol{r_N}}_{\boldsymbol{demos}};\boldsymbol{q}]))
\end{equation}
where $\texttt{Sampling}(\cdot)$ denotes certain sampling strategy and $\texttt{Cat}[\cdot]$ denotes the operation of concatenation.
\paragraph{Scaling Demonstration Number}
Due to restrictions on context window ($2048 \sim 4096$), early studies \citep{ICLfewshot,ICLshortpos} on ICL are limited to few-shot scenarios where they generally observe gains from more demonstrations. 
As the context window expands recently, counterexamples occur. 
\citet{manyICL} finds that the best performance of Gemini 1.5 Pro is achieved under settings where demonstration number is not the maximum one tested in over half of the benchmarks. \citet{controller} discoveries that increasing the number of demonstrations does not necessarily improve model performance across five LLMs. We observe similar phenomena in Figure~\ref{fig:scale}.

\section{Revisiting}
In this section, we explore what impedes LLMs from becoming stable many-shot learners.

\subsection{Approximating ICL as Finetuning}
Since \citet{ICLasFine} derives that ICL is formally equivalent to finetuning, with demonstrations analogous to training samples, we decide to revisit their derivation process below to explore why finetuning satisfies scaling laws \citep{FineScal} while ICL does not.
\paragraph{Finetuning} Let $\boldsymbol{W}_0, \boldsymbol{\Delta W}_{FT} \in \mathbb{R}^{d_{out} \times d_{in}}$ be the initialized parameter matrix and the update matrix, and $\boldsymbol{x} \in \mathbb{R}^{d_{in}}$ be the input representation. The output of certain linear layer optimized by gradient descent can be formulated as follows:
\begin{equation}
    \small
    \boldsymbol{\hat{x}} = \boldsymbol{xW}_0+\boldsymbol{x\Delta W}_{FT}
    \label{equation: finetune}
\end{equation}
\paragraph{ICL} For each attention head of $\boldsymbol{\mathcal{M}}$, let $\boldsymbol{h}_i \in \mathbb{R}^{d_{in}}$ be the representation of the $i$th input token, $\boldsymbol{W}_q, \boldsymbol{W}_k, \boldsymbol{W}_v$ be the projection matrices for computing the queries, keys and values. We denote $\boldsymbol{h}_{i\in \boldsymbol{demos}} \boldsymbol{W}_k $, $\boldsymbol{h}_{i\in \boldsymbol{demos}} \boldsymbol{W}_v $, $\boldsymbol{h}_{i\in \boldsymbol{q}}\boldsymbol{W}_k$, $\boldsymbol{h}_{i\in \boldsymbol{q}}\boldsymbol{W}_v$ as $\boldsymbol{D}_k$, $\boldsymbol{D}_v$, $\boldsymbol{Q}_k$, $\boldsymbol{Q}_v$, respectively. To generate $\boldsymbol{r}$, the output of $\boldsymbol{h_{r}}$ can be derived below:
\begin{equation}
\small
\begin{split}
& \boldsymbol{\hat{h}_{r}} \\
=& \operatorname{Att}(\boldsymbol{h_{r}}\boldsymbol{W}_q,  \operatorname{Cat}[\boldsymbol{D}_k;\boldsymbol{Q}_k ], \operatorname{Cat}[\boldsymbol{D}_v; \boldsymbol{Q}_v]) \\
\approx & \operatorname{LinAtt}(\boldsymbol{h_{r}}\boldsymbol{W}_q , \operatorname{Cat}[\boldsymbol{D}_k; \boldsymbol{Q}_k], \operatorname{Cat}[\boldsymbol{D}_v; \boldsymbol{Q}_v]) \\
= & \boldsymbol{h_{r}}\boldsymbol{W}_q \operatorname{Cat}[\boldsymbol{D}_k; \boldsymbol{Q}_k]^{\top}\left[\begin{array}{c}
\boldsymbol{D}_v \\
\boldsymbol{Q}_v
\end{array}\right] \\
=& \boldsymbol{h_{r}}\boldsymbol{W}_q \boldsymbol{Q}_v \boldsymbol{Q}_k^{\top}+\boldsymbol{h_{r}}\boldsymbol{W}_q \boldsymbol{D}_v\boldsymbol{D}_k^{\top}\\
= &\boldsymbol{h_{r}} \boldsymbol{W}_{ZSL}+\boldsymbol{h_{r}} \Delta \boldsymbol{W}_{ICL} \\
\end{split}
\label{equation: ICL}
\end{equation}
\citet{ICLasFine} approximate the standard attention to linear attention by removing the $\texttt{softmax}$ operation for ease of qualitative analysis. Since $\boldsymbol{h_{r}}\boldsymbol{W}_q \boldsymbol{Q}_v \boldsymbol{Q}_k^{\top}$ is the attention result in the zero-shot learning (ZSL) setting and $\boldsymbol{h_{r}}\boldsymbol{W}_q \boldsymbol{D}_v\boldsymbol{D}_k^{\top}$ is the extra outcome from demonstrations, they are denoted as $\boldsymbol{h_{r}}\boldsymbol{W}_{ZSL}$ and $\boldsymbol{h_{r}}\Delta \boldsymbol{W}_{ICL}$ respectively. Comparing Eq.~\eqref{equation: ICL} with Eq.~\eqref{equation: finetune}, we can understand ICL as finetuning by treating the $\Delta \boldsymbol{W}_{ICL}$ generated from demonstrations as the $\Delta \boldsymbol{W}_{FT}$ generated from training samples. 

\subsection{Ignorance of Attention Competition}
From Eq.~\eqref{equation: ICL} we can further derive as follows:
\begin{equation}
\small
\begin{split}
& \boldsymbol{\hat{h}_{r}} \\
\approx & \underbrace{\operatorname{LinAtt} \left(\boldsymbol{h_{r}}\boldsymbol{W}_q,\boldsymbol{Q}_k,\boldsymbol{Q}_v \right)}_{\text {outcome from }\boldsymbol{q}} + \underbrace{\operatorname{LinAtt} (\boldsymbol{h_{r}}\boldsymbol{W}_q, \boldsymbol{D}_k, \boldsymbol{D}_v }_{\text {outcome from } \boldsymbol{demos}})
\end{split}
\label{equation: ICL_further}
\end{equation}
which means that the existence of demonstrations does not affect the outcome from $q$. 
However, when we no longer approximate standard attention as linear attention, we arrive at the opposite conclusion:
\begin{equation}
\small
\begin{aligned}
& \boldsymbol{\hat{h}_{r}} \\
=& \operatorname{Att}(\boldsymbol{h_{r}}\boldsymbol{W}_q,  \operatorname{Cat}[\boldsymbol{D}_k;\boldsymbol{Q}_k ], \operatorname{Cat}[\boldsymbol{D}_v; \boldsymbol{Q}_v]) \\
=&\operatorname{softmax}(\boldsymbol{h_{r}}\boldsymbol{W}_q \operatorname{Cat}[\boldsymbol{D}_k; \boldsymbol{Q}_k]^{\top})
\left[\begin{array}{c}
\boldsymbol{D}_v \\
\boldsymbol{Q}_v
\end{array}\right] \\
=&(1-\lambda(\boldsymbol{h_{r}})) \operatorname{softmax}(\boldsymbol{h_{r}}\boldsymbol{W}_q \boldsymbol{Q}_k^{\top}) \boldsymbol{Q}_v\\
&+\lambda(\boldsymbol{h_{r}}) \operatorname{softmax}(\boldsymbol{h_{r}}\boldsymbol{W}_q \boldsymbol{D}_k^{\top}) \boldsymbol{D}_v \\
=&(1-\lambda(\boldsymbol{h_{r}})) \underbrace{\operatorname{Att}\left(\boldsymbol{h_{r}}\boldsymbol{W}_q, \boldsymbol{Q}_k, \boldsymbol{Q}_v\right)}_{\text {outcome from }\boldsymbol{q}}\\
&+\lambda(\boldsymbol{h_{r}}) \underbrace{\operatorname{Att}\left(\boldsymbol{h_{r}}\boldsymbol{W}_q, \boldsymbol{D}_k, \boldsymbol{D}_v\right)}_{\text {outcome from } \boldsymbol{demos}},
\end{aligned}
\label{equation: ICL_new}
\end{equation}
where:
\begin{equation}
\small
\lambda(\boldsymbol{h_{r}})=\frac{\sum_i \exp \left(\boldsymbol{h_{r}}\boldsymbol{W}_q \boldsymbol{D}_k^{\top}\right)_i}{\sum_i \exp \left(\boldsymbol{h_{r}}\boldsymbol{W}_q \boldsymbol{D}_k^{\top}\right)_i+\sum_j \exp \left(\boldsymbol{h_{r}}\boldsymbol{W}_q \boldsymbol{Q}_k^{\top} \right)_j}
\label{equation: ICL_lambda}
\end{equation}
With the existence of $\lambda(\boldsymbol{h_{r}})$ in Eq.~\eqref{equation: ICL_new}, an increase in the number of demonstrations will lead to a larger $\lambda(\boldsymbol{h_{r}})$, thereby decreasing the model attention towards $q$. 
At the same time, ICL does not necessarily adhere to the scaling law as it is no longer formally equivalent to finetuning.
\textbf{Therefore, we hypothesize that more demonstrations can divert model attention from the key contents (query), leading to possible performance decrease.}

\begin{figure}[b]
\centering
\includegraphics[width=0.48\textwidth]{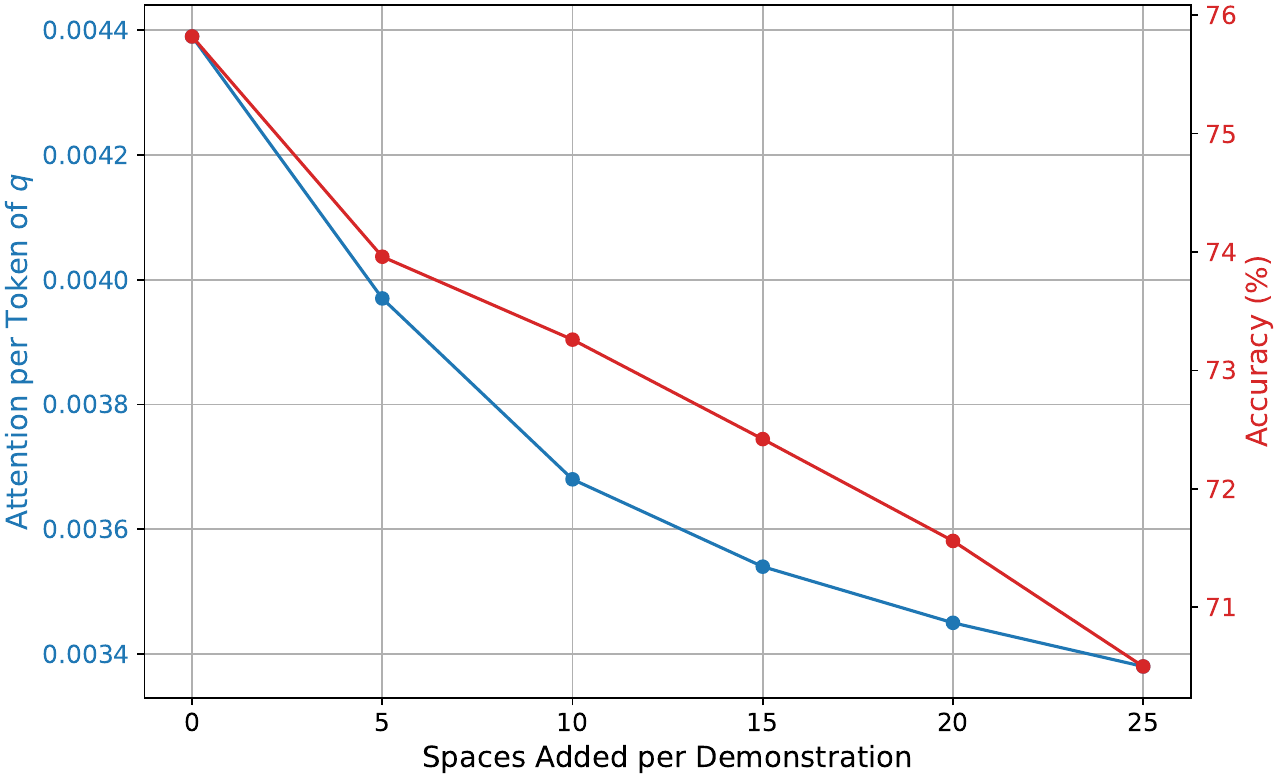} 
\caption{Accuracy and attention of \textsc{longchat-7b-v1.5-32k} with varying number of spaces added per demonstration. Demonstration number is set as 100.}
\label{fig:pad}
\end{figure}

\begin{figure*}[t]
\centering
\includegraphics[width=0.96\textwidth]{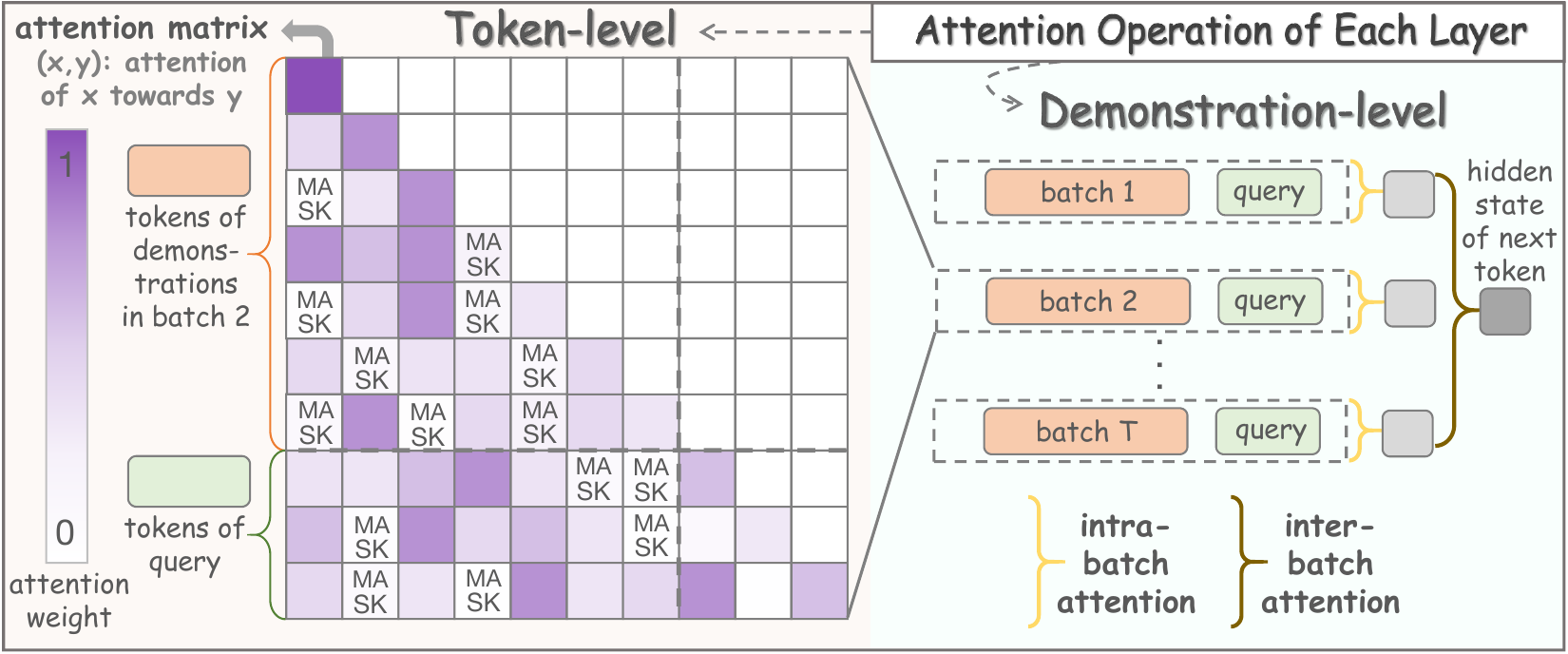} 
\caption{Overall illustration of \textsc{FocusICL}.}
\label{fig:main}
\end{figure*}

\subsection{Experimental Evidence for Hypothesis}
\label{sec:pre_exp}
To validate our hypothesis, we first investigate whether the model attention towards the query decreases with the increase of demonstration number. 
To avoid potentially unreliable results caused by data contamination \citep{datacon}, our exploratory experiments are conducted with longchat-7b-v1.5 \citep{longchat} (32k context window) on the proposed \textsc{CountA} benchmark (See details in \S\ref{sec:benchmark}), which requires the model to \textbf{Count} the number of character ‘\textbf{A}' in the five candidates.  
As shown in Figure~\ref{fig:abs}, the average attention weight of model towards each token in the query decreases by scaling up the demonstration number, corresponding to Eq.~\eqref{equation: ICL_new}.

We further explore how the model's lack of attention towards the query affects the quality of the response. Specifically, we add several blank spaces at the end of each demonstration. This format maintains the ICL paradigm and the meaningless blank spaces will not introduce additional information. As shown in Figure~\ref{fig:pad}, we find that more blank spaces disperse the model attention towards the query similar to the demonstrations, which in turn leads to a decline in accuracy. Based on the experiments above, we have confirmed our hypothesis.

\section{Methodology}

To mitigate the impact of LLMs' attention being dispersed by many-shot demonstrations, we propose \textsc{FocusICL}. The core idea behind \textsc{FocusICL} is to allocate model attention to more important contents at token-level by triviality filtering (\S\ref{sec:tri}) and at demonstration-level by hierarchical attention (\S\ref{sec:hir}), as shown in Figure~\ref{fig:main}.
\subsection{Triviality Filtering}
\label{sec:tri}
Humans benefit from selectively ignoring irrelevant parts (trivialities) of demonstrations to avoid attention dispersion. 
In contrast, the standard attention mechanism of LLMs fails to completely ignore (assign zero attention weight to) trivialities and leverage the prior that the tokens of query are generally important, 
for which we propose triviality filtering operation.
To predict response $\boldsymbol{r}$ for given query $\boldsymbol{q}$, in each attention layer, we first calculate the attention scores $\boldsymbol{s}$ as follows:
\begin{equation}
\small
\boldsymbol{s}=\boldsymbol{h_{r}}\boldsymbol{W}_q  \operatorname{Cat}[\boldsymbol{D}_k;\boldsymbol{Q}_k ]^{\top}
\end{equation}
Instead of directly applying $\texttt{softmax}$ on $\boldsymbol{s}$ like standard attention operation, we filter the trivialities in the demonstrations according to a pre-set threshold $p$ in advance as follows:
\begin{equation}
\small
\begin{gathered}
    \texttt{index} = \texttt{arg}\{\texttt{index}|\texttt{count}(\boldsymbol{s} \leq \boldsymbol{s}_\texttt{index})=p\times|\boldsymbol{s}|\} \\
    \texttt{mask}(\boldsymbol{s}) = \begin{cases}
-\texttt{INF},\ \boldsymbol{s}_i \leq \boldsymbol{s}_\texttt{index}\ \texttt{and}\ i \in \boldsymbol{demos}
 \\
0,\ \texttt{else}
\end{cases} \\
\boldsymbol{\hat{h}_{r}}=\texttt{softmax}(\boldsymbol{s}+\texttt{mask}(\boldsymbol{s}))\operatorname{Cat}[\boldsymbol{D}_v;\boldsymbol{Q}_v ]
\end{gathered}
\end{equation}
where $\boldsymbol{\hat{h}_{r}}$ is the outcome of $\boldsymbol{h_{r}}$. 
By applying triviality filtering operation, useless parts of demonstrations are assigned zero attention weights thus
LLMs can focus on leveraging relevant contents of the demonstrations to solve the current query.
To achieve a broad impact, apart from $\boldsymbol{r}$, we also apply triviality filtering operation on tokens belong to responses of demonstrations by autoregressively treating $\{(\boldsymbol{q}_i,\boldsymbol{r}_i)\}_{i=1}^{k-1}$ as demonstrations of $(\boldsymbol{q}_k,\boldsymbol{r}_k), k \in [2,N]$.


\subsection{Hierarchical Attention}
\label{sec:hir}
\begin{figure}[t]
\centering
\includegraphics[width=0.48\textwidth]{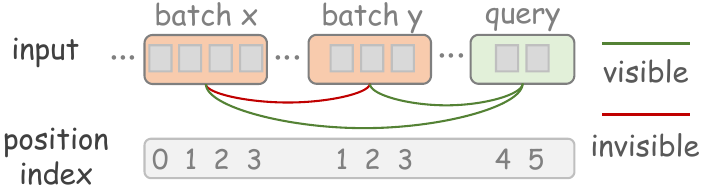} 
\caption{Input details of \textsc{FocusICL}.}
\label{fig:format}
\end{figure}

When there are numerous examples, humans draw inspirations for problem-solving from different examples separately and then integrate the insights to avoid distracting attention by focusing on too many examples simultaneously.
Motivated by this, we introduce hierarchical attention mechanism for LLMs to learn from many-shot demonstrations while focusing on current query.
We first split the demonstrations into $T$ batches, where each one comprises $B$ consecutive demonstrations.
Without editing the token order, we change the position indexes to ensure that each batch is logically adjacent to the query (Figure~\ref{fig:format}). 
To ensure that batches are mutually invisible to each other, we use a mask matrix, allowing us to parallelly apply intra-batch attention within each batch $i$ and query as follows:
\begin{equation}
\small
\boldsymbol{\hat{h}_{r}^i},\ \boldsymbol{s^i}=\texttt{TrivialityFiltering Att}(\boldsymbol{h}_{j \in batch_i \cup \boldsymbol{q} })
\end{equation}
By controlling the batch size $B$, we can ensure that the model maintains enough attention towards the query within each batch.
To further integrate insights from different batches, we conduct inter-batch attention as follows:
\begin{equation}
\small
\boldsymbol{\hat{h}_{r}}=\sum_{i=1}^{T} \boldsymbol{\hat{h}_{r}^i}\times \frac{\sum_{j} e^{s^i_j}}{\sum_{k}\sum_{j} e^{s^k_j}}
\end{equation}
The sum of the attention scores for all tokens within each batch can reflect the amount of useful information contained in that batch for the current query. Based on this, we calculate the weighted sum of $\boldsymbol{\hat{h}_{r}^i}$ to attain the final output of the attention layer.


\subsection{Hyperparameter Searching}
To efficiently find suitable values of filtering threshold $p$ and batch size $B$ for different LLMs and tasks, we propose a hyperparameter searching strategy as shown in Algorithm~\ref{alg:1}. 
\begin{algorithm}
\small
    \caption{Hyperparameter Searching.}\label{alg:1}
        \begin{algorithmic}[1]
\Require Candidate filtering threshold set $\boldsymbol{\mathcal{S}_{p}}$,  LLM $\boldsymbol{\mathcal{M}}$ \newline
Demonstration set $\boldsymbol{\mathcal{S}_{demos}}$, Demonstration number $N$
\Ensure Suitable filtering threshold $p$ and batch size $B$
\State $D(p,i) \leftarrow 0$ for $p \in \mathcal{S}_{p}, i \in [0,N-1]$
\For{$p \in \boldsymbol{\mathcal{S}_{p}}$}:
    \For{$i\gets 1, 5$}:
        \State $\boldsymbol{\mathcal{S}_{1:N}} \leftarrow$ \texttt{RandomSelect}($\boldsymbol{\mathcal{S}_{demos}},N$)
        \State \# calculate average $ppl$ of responses in $\mathcal{S}_{1:N}$
        \State $\boldsymbol{ppl_{1:N}} \leftarrow \boldsymbol{\mathcal{M}}(\texttt{ICLFormat}(\boldsymbol{\mathcal{S}_{1:N}}))$
            \State $D(p,j-1) \leftarrow D(p,j-1)+\boldsymbol{ppl_{j}}$ for $j\in [1,N]$
    \EndFor
    \State $D(p,i) \leftarrow D(p,i)+D(p,i+1) $ for $i\in [0,N-2]$
    \State $\Bar{D}(p,i) \leftarrow D(p,i)-D(p,i-2) $ for $i\in [2,N-2]$
\EndFor
\State $p\leftarrow \texttt{argmin}(p|\texttt{sum}(D(p)))$
\State $B\leftarrow \texttt{argmin}(i|\Bar{D}(p,i)>0)$
\end{algorithmic}
\end{algorithm}
By treating $q_i$ as current query and $\mathcal{S}_{1:i-1}$ as demonstrations, the model perplexity \footnote{We don't use accuracy because the accuracy obtained under teacher forcing will overestimate the model performance.} ($ppl$) of $r_i$ can reflect the LLMs' capability when demonstration number is $i-1$ (lower $ppl$ indicates better performance). Thus, we choose the $p$ that yields the lowest average $ppl$ and $B$ that first leads an increasing trend in $ppl$ as our hyperparameter choices.
We generally set $\boldsymbol{\mathcal{S}_{p}}$ as $[0,0.1,0.2,0.3,0.4]$ and run each setting 5 times to stabilize the results, resulting in a total of 25 inference overhead for hyperparameter searching, which is relatively low compared with the thousands of evaluation samples.

\section{Experiments}
Centered around \textsc{FocusICL}, we will empirically demonstrate its performance on different LLMs and tasks in \S\ref{sec:overall}, verify whether it can help LLMs scale well with demonstration number in \S\ref{sec:scale}, and delve into its working mechanism in \S\ref{sec:analysis}. We also investigate the choice of hyperparameters in Appendix~\S\ref{appendix_hyperdis}.
\subsection{Experimental Settings}
\paragraph{Benchmarks} We conduct experiments on the following benchmarks:
\label{sec:benchmark}
\begin{itemize}[leftmargin=20pt]
\setlength{\itemsep}{0pt}
\setlength{\parsep}{0pt}
\setlength{\parskip}{0pt}
\item \textbf{CSQA} \citep{csqa} is a high-quality benchmark for commonsense reasoning task.
\item \textbf{PIQA} \citep{piqa} concentrates on testing physical commonsense answering ability.
\item \textbf{CountA} is our proposed benchmark to avoid the impact of data contamination \citep{datacon}, making experimental results more comprehensive and reliable. It requires the model to count the number of character 'A' in the five candidates.
\item \textbf{ARC} \citep{arc} includes questions that require extensive knowledge and reasoning to answer. 
\item \textbf{GSM8K} \citep{gsm8k} serves as a testbed for evaluating multi-step mathematical reasoning (chain-of-thought) ability.
\end{itemize}
We evaluate the LLMs on the test set of the datasets above and use the training set as the demonstration candidate set $\boldsymbol{\mathcal{S}_{demos}}$. 
\paragraph{Baselines} 
\begin{itemize}[leftmargin=20pt]
\setlength{\itemsep}{0pt}
\setlength{\parsep}{0pt}
\setlength{\parskip}{0pt}
\item \textbf{\textsc{ICL}.} We use a unified ICL \citep{ICLfewshot} input format for all the methods for fair comparisons, as shown in Appendix~\S\ref{prompt}.
\item \textbf{\textsc{EarlyStop}.} \citet{controller} proposes to pick the optimal demonstration number according to the performance on a validation set.
\item \textbf{\textsc{StructICL}.} \citet{structure} share a similar idea with us of dividing demonstrations into batches. Differently, their designs focus on extending available context length.
\end{itemize}

\paragraph{Details} We conduct experiments with three widely used long-context LLMs: \textsc{longchat-7b-v1.5-32k} \citep{longchat}, \textsc{vicuna-7b-v1.5-16k} \citep{vicuna} and \textsc{Llama-3-8B-Instruct} \citep{llama3}. We choose the maximum available number of demonstrations for evaluation based on the 40 GB memory of the A100 GPU (Table~\ref{appendixtab: demonum}). The hyper parameter searching results are listed in Table~\ref{appendixtab: hyper}.  We use random sampling decoding strategy (T=0.1) and report the outcomes averaged over 5 runs (randomly selecting demonstrations) for credible results. 

\begin{table}[t]
    \renewcommand\arraystretch{1.3}
    \small
    \centering
    \setlength{\tabcolsep}{0.13em} 
    \begin{tabular}{lcccccc}
    \toprule
    \textbf{Method}&\textbf{CSQA} &\textbf{PIQA} &\textbf{CountA}  & \textbf{ARC}&\textbf{GSM8K}&\textbf{Avg.}\\
    \midrule
     \textsc{ICL}&47.58&57.42&79.04&62.43&9.93&51.28\\
     \cdashline{1-7}
     \textsc{EarlyStop}&47.89&57.44&81.28&62.43&11.14&52.04\\
     \textsc{StructICL}&50.25&59.02&86.77&64.05&11.25&54.27\\
     \cdashline{1-7}
     \textsc{Triviality}&48.97&58.65&85.68&63.13&11.00&53.49\\
     \textsc{\textbf{FocusICL}}&\textbf{50.70}&\textbf{60.83}&\textbf{91.94}&\textbf{64.55}&\textbf{12.28}&\textbf{56.06}\\
    \bottomrule
    \end{tabular}
    \caption{Accuracy (\%) of \textsc{longchat-7b-v1.5-32k} with compared methods across benchmarks.}
    \label{tab: Main results1}
\end{table}

\begin{table}[t]
    \renewcommand\arraystretch{1.3}
    \small
    \centering
    \setlength{\tabcolsep}{0.13em} 
    \begin{tabular}{lcccccc}
    \toprule
    \textbf{Method}&\textbf{CSQA} &\textbf{PIQA} &\textbf{CountA}  & \textbf{ARC}&\textbf{GSM8K}&\textbf{Avg.}\\
    \midrule
     \textsc{ICL}&60.72&60.09&82.20&77.11&16.30&59.23\\
     \cdashline{1-7}
     \textsc{EarlyStop}&61.36&60.20&82.20&78.14&17.44&59.87\\
     \textsc{StructICL}&61.44&61.81&84.78&78.05&17.12&60.64\\
     \cdashline{1-7}
     \textsc{Triviality}&61.51&61.03&84.43&77.78&17.36&60.42\\
    \textsc{\textbf{FocusICL}}&\textbf{62.57}&\textbf{67.88}&\textbf{85.13}&\textbf{78.51}&\textbf{17.74}&\textbf{62.37}\\
    \bottomrule
    \end{tabular}
    \caption{Accuracy (\%) of \textsc{vicuna-7b-v1.5-16k} with compared methods across benchmarks.}
    \label{tab: Main results2}
\end{table}

\begin{table}[t]
    \renewcommand\arraystretch{1.3}
    \small
    \centering
    \setlength{\tabcolsep}{0.13em} 
    \begin{tabular}{lcccccc}
    \toprule
    \textbf{Method}&\textbf{CSQA} &\textbf{PIQA} &\textbf{CountA}  & \textbf{ARC}&\textbf{GSM8K}&\textbf{Avg.}\\
    \midrule
     \textsc{ICL}&74.90&75.86&98.10&90.00&66.64&81.10\\
     \cdashline{1-7}
     \textsc{EarlyStop}&75.54&77.09&98.10&90.47&71.21&82.48\\
     \textsc{StructICL}&75.12&77.05&98.16&90.70&69.43&82.09\\
     \cdashline{1-7}
     \textsc{Triviality}&75.25&76.38&98.22&90.40&68.03&81.56\\
    \textsc{\textbf{FocusICL}}&\textbf{76.00}&\textbf{78.29}&\textbf{98.34}&\textbf{91.02}&\textbf{71.89}&\textbf{83.11}\\
    \bottomrule
    \end{tabular}
    \caption{Accuracy (\%) of \textsc{Llama-3-8B-Instruct} with compared methods across benchmarks.}
    \label{tab: Main results3}
\end{table}

\subsection{Main Results}
\label{sec:overall}
Our main experimental results are presented in Tables~\ref{tab: Main results1}, \ref{tab: Main results2}, and \ref{tab: Main results3}. The compared methods exhibit similar performance trends across different LLMs. 
\paragraph{Baselines} Under most settings, \textsc{EarlyStop} outperforms \textsc{ICL}, consistent with the observations of \citet{manyICL} and \citet{controller} that more demonstrations does not necessarily lead to better performance. Compared to \textsc{EarlyStop} which avoids the negative impact of attention dispersion by not introducing more demonstrations, \textsc{StructICL} leverages all the given demonstrations through structured input to achieve slightly better performance.
\paragraph{Ours} However, due to the lack of insights into the reasons behind performance degradation of \textsc{ICL} with more demonstrations, the baselines fail to maintain the model attention on critical input parts while fully leveraging all demonstrations. In contrast, by introducing triviality filtering operation and hierarchical attention mechanism to achieve the above vision, \textsc{FocusICL} outperforms the compared baselines, achieving an average of 5.2\% (3.31 points) performance improvement over \textsc{ICL} across three LLMs. The results of the T-test also indicate that \textsc{FocusICL} is significantly superior to baselines, with a p-value less than 0.05. This validates the effectiveness and generalizability of \textsc{FocusICL}.
\paragraph{Ablations} We also report the performance of only performing triviality filtering operation as an ablation study. The results show that \textsc{FocusICL} benefits 1.29 points improvement from the triviality filtering operation and 2.02 points improvement from the hierarchical attention mechanism.
\paragraph{Efficiency} By performing hierarchical attention mechanism, demonstrations between different batches does not need direct interactions, which can save a significant amount of inference overhead. 
Assuming each demonstration has an average of $L$ tokens, the overhead of attention operation between $N$ demonstrations for ICL is:
\begin{equation}
\small
    Cost_{\textsc{ICL}} = N^2L^2\times \Delta
\end{equation}
where $\Delta$ denotes a computational cost unit.
The overhead for \textsc{FocusICL} with batch size as $B$ is:
\begin{equation}
\small
\begin{aligned}
    Cost_{\textsc{FocusICL}} & = \frac{N}{B}(BL)^2\times \Delta \\
    & = NBL^2 \times \Delta
\end{aligned}
\end{equation}
Therefore, the overhead ratio of \textsc{FocusICL} to \textsc{ICL} in encoding demonstrations is $B:N$ ($N$ is generally several times larger than $B$), while the overhead in other aspects is roughly the same. This demonstrates the efficiency of \textsc{FocusICL}.

\begin{figure*}[t]
\centering
\includegraphics[width=1\textwidth]{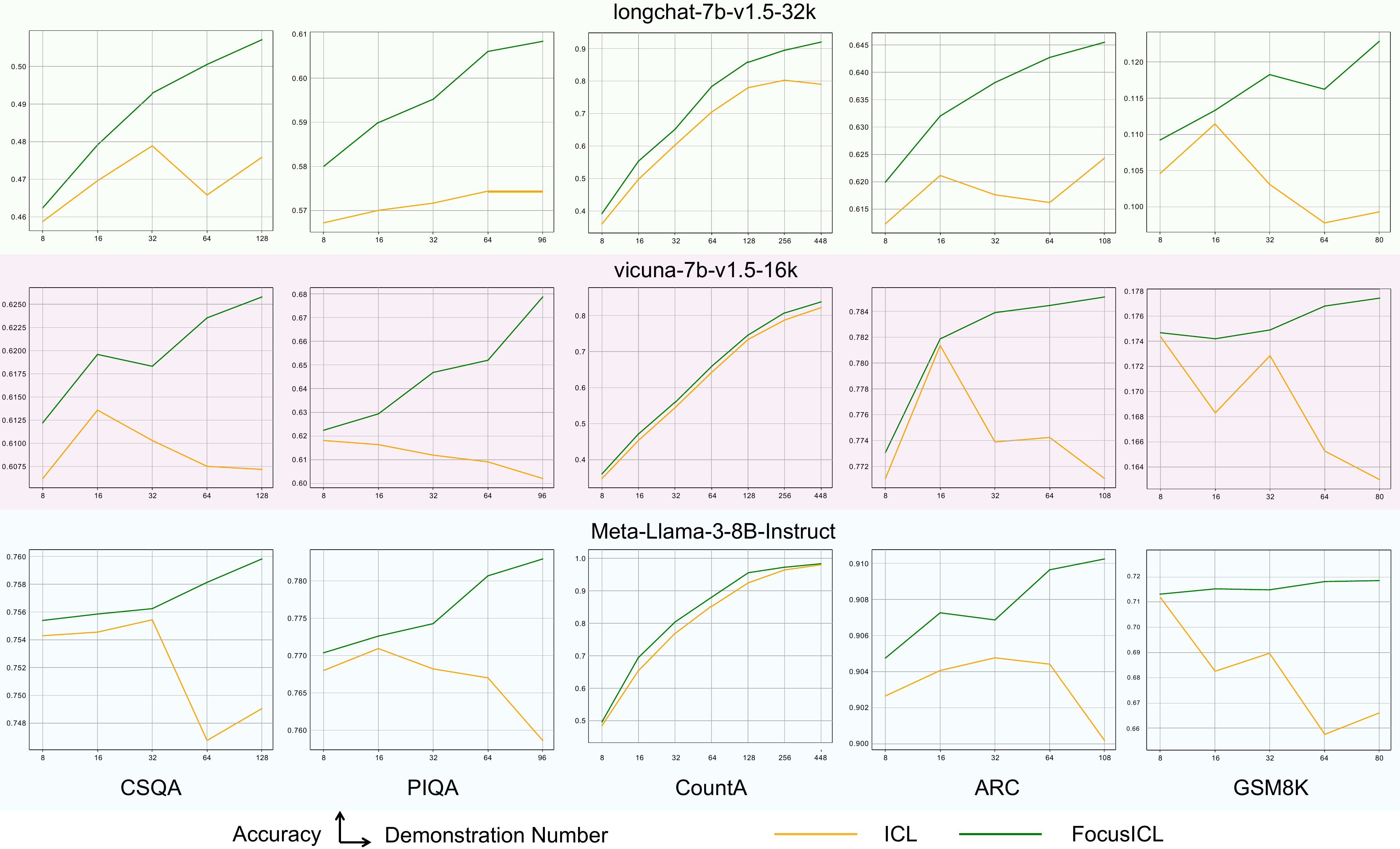} 
\caption{\textsc{FocusICL} helps different LLMs scale well with many-shot demonstrations compared with \textsc{ICL}.}
\label{fig:scale}
\end{figure*}

\subsection{Scaling with More Demonstrations}
\label{sec:scale}
The recent significant advancements in LLMs mainly stem from scaling up in dimensions of model size and training data size. 
However, given the limitations of computation resource and data production speed, we are in eager need of exploring other potential scaling dimensions to continuously enhance the performance of LLMs. 
As shown in Figure~\ref{fig:scale}, the demonstration number is not a stable scaling dimension when applying ICL,
as the performance can sometimes exhibit an inverse-scaling phenomenon with more demonstrations.
In contrast, \textsc{FocusICL} enables LLMs to become stable many-shot learners by directing their attention to important contents, thereby achieving good scalability in the dimension of demonstration number.

It should be noted that we find the advantage of \textsc{FocusICL} over ICL continues to grow as the number of demonstrations increases. This means that if we have more resources to conduct experiments with more demonstrations, the advantage of \textsc{FocusICL} over ICL can be larger.

\subsection{Working Mechanism of \textsc{FocusICL}}
\label{sec:analysis}
To gain a deeper understanding of the working mechanism of \textsc{FocusICL}, we explore it from aspects of attention and hidden state distributions, following the experimental settings in \S\ref{sec:pre_exp}.

\begin{figure}[t]
\centering
\includegraphics[width=0.48\textwidth]{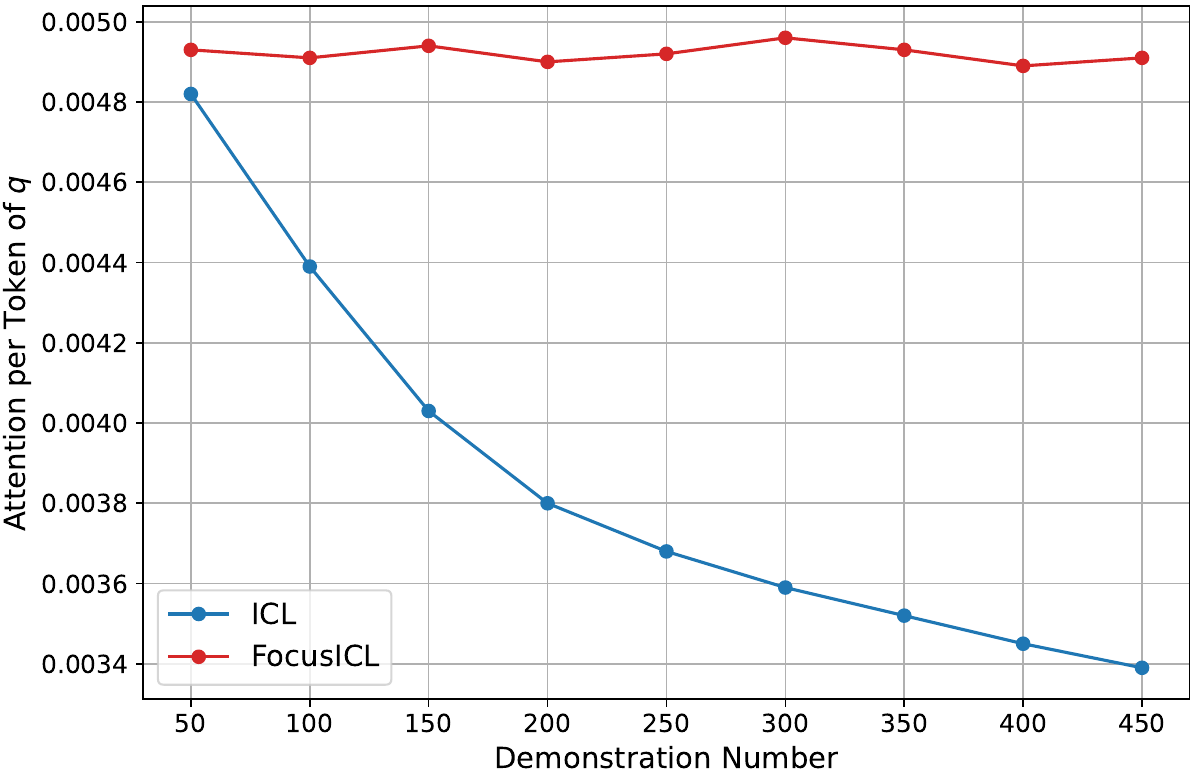} 
\caption{Average model attention towards token of $q$ with varying demonstration numbers.}
\label{fig:attention}
\end{figure}
\paragraph{Attention Distribution}
The primary purpose of \textsc{FocusICL} is to prevent the model attention from being scattered by the increased demonstrations, thereby ensuring a proper understanding of key contents. 
Therefore, we observe the attention weights allocated by the model towards the query as the number of demonstrations increases. As shown in Figure~\ref{fig:attention}, by ignoring unimportant parts of the demonstrations and introducing the hierarchical attention mechanism, \textsc{FocusICL}  consistently maintains sufficient attention towards the query.

\begin{figure}[!htb]
    \centering
    \subfigure{\includegraphics[width=0.99\hsize, height=0.66\hsize]{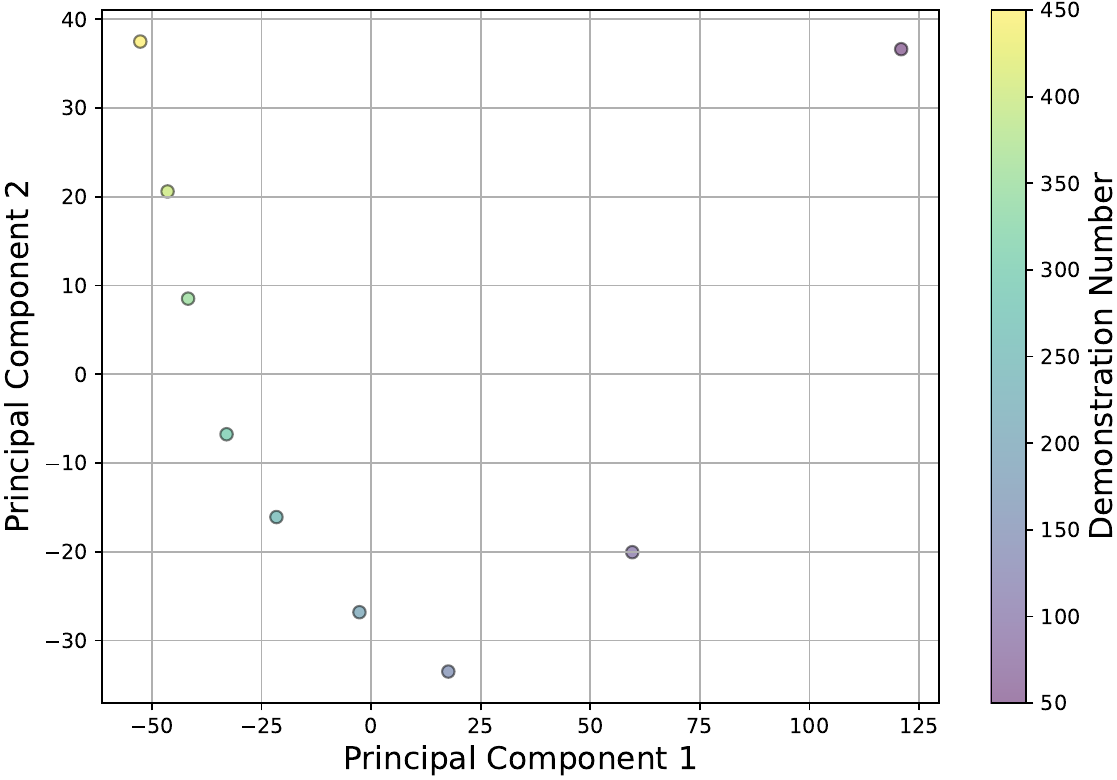}\label{fig: sub_figure1}} 
    \vspace{-3mm}
    \subfigure{\includegraphics[width=0.99\hsize, height=0.66\hsize]{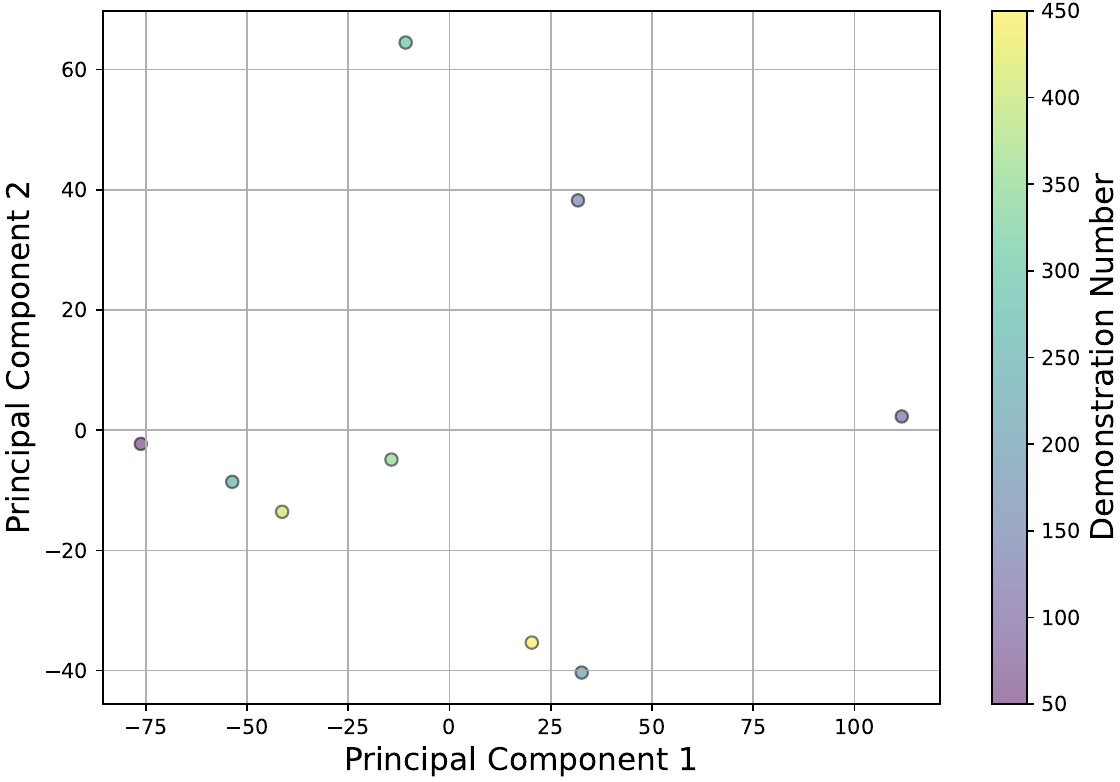}\label{fig: sub_figure2}} 
    \caption{The PCA distribution results of the hidden states of the last input token from the penultimate layer of ICL (above) and \textsc{FocusICL} (below) with varying numbers of demonstrations.}
    \setlength{\belowcaptionskip}{-5pt}
\label{fig:pca}
\end{figure}
\paragraph{Hidden States Distribution}
We further investigate the distribution of the hidden states of the last input token at the penultimate model layer through Principal Component Analysis (PCA). 
Intuitively, the distribution of the hidden states of the last token mainly depends on the current problem to be solved and should be independent of the demonstration number.
However, as shown in Figure~\ref{fig:pca}, we find that the hidden states of ICL change systematically with an increasing number of demonstrations, whereas \textsc{FocusICL} does not exhibit such behavior.
We think that the systematic decline in attention towards the query in ICL with an increasing number of demonstrations continuously affects the hidden states during response generation, thereby impacting the quality of the generated response. In contrast, \textsc{FocusICL} avoids this issue by maintaining sufficient attention to the query as shown above, ultimately benefiting from more demonstrations.
\subsection{Further Discussion}
Based on our existing insights and experimental results, we attempt to understand the divergent phenomena of ICL observed in previous studies where more demonstrations sometimes lead to better performance, while sometimes the opposite occurs. 
We think the main reason leading to the above phenomena comes from the double-edged sword effect of learning from more demonstrations: on the one hand, they can help the model better understand the task and increase the likelihood of finding useful knowledge; on the other hand, they might also distract the model, leading to insufficient attention and understanding of current query.
We consider that two aspects can influence the balance between the two effects:
\paragraph{Weak models require more demonstrations to understand the task.}
As shown in Figure~\ref{fig:scale}, we observe that the optimal number of demonstrations for \textsc{longchat-7b-v1.5-32k} is greater compared to the other two models across most benchmarks. 
Considering that its performance is also the worst, we believe the reason for the aforementioned situation is that weaker models require more demonstrations to help them better understand the task.
\paragraph{More demonstrations are needed when they have a closer relationship.}
We also notice that the LLMs are more demonstration-hungry on CountA compared to other benchmarks as shown in Figure~\ref{fig:scale}. 
We analyze that the correlation between samples in other benchmarks is relatively weak, 
and even a single demonstration is sufficient to clarify the task format. In contrast, the demonstrations in CountA are closely related, collectively determining what the task definition is. In this scenario, LLMs cannot discern the complete task information if only given a few demonstrations.
To sum up, when the samples are closely related, the model needs more demonstrations to analyze the correlations among them, so as to better understand and complete the task.


\section{Conclusions}
Noticing that the performance of LLMs under many-shot ICL does not consistently improve with more demonstrations, we analyze and validate the underlying reason as follows: more demonstrations can disperse the model attention to critical contents, resulting in an insufficient understanding of the query. Inspired by how humans learn from examples, we propose a training-free method \textsc{FocusICL}, which conducts triviality filtering at token-level and hierarchical attention at demonstration-level to rationally allocate model attention in each layer. Comprehensive experiments indicate that focused LLMs are stable many-shot learners, making demonstration number a possible scaling dimension for LLM-based AGI.

\section*{Limitations}
From an objective perspective , we think there are two main limitations of this paper:
\begin{enumerate}

\item Although we have extended the demonstration number to nearly or even beyond 100, due to computational resource limitations, we are unable to conduct experiments with larger demonstration numbers. We will further verify the applicability of \textsc{FocusICL} with larger demonstration numbers in the future.

\item This work primarily discusses LLMs that apply the standard transformer decoder architecture. We look forward to further exploring the scaling performance with the demonstration number and the applicability of \textsc{FocusICL} on other variants of LLMs, such as the encoder-decoder architecture and sliding window attention, in the future.

\end{enumerate}

\section*{Ethics Statement}
All of the datasets used in this study were publicly available, and no annotators were employed for our data collection. We confirm that the datasets we used did not contain any harmful content and was consistent with their intended use (research). We have cited the datasets and relevant works used in this study.

\bibliography{acl_latex}

\appendix
\clearpage
\appendix
\label{sec:appendix}

\section{Additional Experimental Results}
\subsection{Hyperparameters}
\label{appendix_hyperdis}
To investigate the influence of hyperparameters, we report the results of \textsc{longchat-7b-v1.5-32k} on GSM8K benchmark with varying hyperparameter settings. 
\paragraph{Filtering Threshold} As shown in Table~\ref{appendixtab: Main results1}, with the increase of filtering threshold $p$, the model's performance first improves and then declines. This is because, when $p$ is relatively small, the model benefits from ignoring unimportant content and focusing its attention on more beneficial parts. However, when $p$ becomes larger, the model might overlook potentially useful information in the demonstrations, leading to a decrease in performance.
\paragraph{Batch Size} As shown in Table~\ref{appendixtab: Main results2}, a similar inverted U-shaped curve phenomenon occurs when scaling the batch size while maintaining the overall demonstration number fixed. 
As the batch size decreases from 80, the model attention to the query continues to increase, which can lead to a certain improvement in model performance. However, when the batch size is too small, the model may fail to fully understand the task definition due to excessive lack of interaction between demonstrations, consistent with the findings of \citet{ICLscale}.

Luckily, through our proposed hyperparameter searching strategy, we can efficiently attain suitable hyperparameters for the given tasks and LLMs.

\subsection{Further Analyses of \textsc{Triviality}}
When we identify tokens that are unhelpful for answering the current query through attention, \textsc{Triviality} directly masks them to prevent the model's attention from being distracted. Another more intuitive approach is to filter out demonstrations with minimal attention. We compared these two methods, and the results are shown in the Table~\ref{appendixtab: triviality1}. It can be seen that \textsc{Triviality}, which operates at a finer granularity at the token level, achieves better results.

Additionally, we conducted the following experiments to further validate the motivation that tokens with low attention are unimportant and should be masked. We set the following settings below on CountA with LONGCHAT-7B-V1.5-32K:
\begin{itemize}[leftmargin=20pt]
\setlength{\itemsep}{0pt}
\setlength{\parsep}{0pt}
\setlength{\parskip}{0pt}
\item \textbf{No Masking.}
\item \textbf{Masking 40\% of tokens with the lowest attention.} 
\item \textbf{Masking 40\% of tokens with the highest attention.} 
\item \textbf{Randomly masking 40\% of tokens.} 
\end{itemize}
The experimental results in Table~\ref{appendixtab: triviality2} demonstrate the following: compared to No Masking, randomly masking reduces accuracy from 79.04\% to 35.00\%. Masking high-attention tokens leads the model to repeatedly output the word 'nobody', indicating a loss of problem-solving ability. Conversely, masking low-attention tokens significantly improves performance.

To further analyze the underlying reasons, we calculated the model's perplexity across different settings. We found that random masking and masking high-attention tokens significantly increase model perplexity, likely due to the loss of critical token information. In contrast, masking low-attention tokens decreases model perplexity, indicating that filtering trivial tokens based on posterior attention information helps the model perform tasks more confidently.

\begin{table}[ht]
    \renewcommand\arraystretch{1.3}
    \small
    \centering
    \setlength{\tabcolsep}{0.13em} 
    \begin{tabular}{ccccc}
    \toprule
    \textbf{Method}&\textsc{ICL} &\textsc{ICL-drop} &\textsc{Triviality}  & \textsc{FocusICL}\\
    \midrule
     Accuracy&9.93&10.79&11.00&12.28\\
    \bottomrule
    \end{tabular}
    \caption{Accuracy (\%) of different methods on GSM8K with LONGCHAT-7B-V1.5-32K. ICL-drop indicates the ICL method with dropping the 10 demonstrations with lowest average attention weights.}
    \label{appendixtab: triviality1}
\end{table}

\begin{table}[ht]
    \renewcommand\arraystretch{1.3}
    \small
    \centering
    \setlength{\tabcolsep}{0.13em} 
    \begin{tabular}{ccc}
    \toprule
    \textbf{Method}&\textbf{Accuracy} &\textbf{PPL}\\
    \midrule
     No Masking&79.04	&0.610\\
     Mask Low-attention Tokens	&85.68	&0.572\\
     Mask High-attention Tokens&0.00	&10.921\\
     Random Masking&35.00	&1.636\\
    \bottomrule
    \end{tabular}
    \caption{Accuracy (\%) of different methods on GSM8K with LONGCHAT-7B-V1.5-32K.}
    \label{appendixtab: triviality2}
\end{table}

\subsection{\textsc{FocusICL} with Demonstrations Retrieval}
Previous research \citep{demoret1,demoret2,demoret3} have shown that selecting demonstrations relevant to the current query can enhance the performance of ICL. We investigated whether combining \textsc{FocusICL} with demonstration retrieval could yield better results. For simplicity, we used BERT embeddings rather than other complex retrieval methods \citep{ret1} to retrieve the most relevant demonstrations. We then compared the experimental results using both ICL and FocusICL, as shown in Table~\ref{appendixtab: ret}. Retrieving relevant demonstrations resulted in a 1.13\% improvement for ICL and a 1.53\% enhancement for FocusICL. This improvement is likely attributed to the hierarchical attention mechanism's ability to more effectively utilize demonstrations with substantial informative content.
\begin{table}[ht]
    \renewcommand\arraystretch{1.3}
    \small
    \centering
    \setlength{\tabcolsep}{0.13em} 
    \begin{tabular}{ccc}
    \toprule
    \textbf{Method}&\ \ \ ICL &\textsc{FocusICL}\\
    \midrule
     Random Demonstrations&\ \ \ 47.58	&50.70\\
     Relevant Demonstrations	&\ \ \ 48.71	&52.23\\
    \bottomrule
    \end{tabular}
    \caption{Accuracy (\%) of different methods on CSQA with LONGCHAT-7B-V1.5-32K.}
    \label{appendixtab: ret}
\end{table}

\section{Derivation Details}
The derivation details of Equation~\eqref{equation: ICL_new} are as follows:
\begin{equation}
\small
\begin{aligned}
& \text { output } \\
=& \operatorname{Att}(\boldsymbol{h_{r}}\boldsymbol{W}_q,  \operatorname{Cat}[\boldsymbol{D}_k;\boldsymbol{Q}_k ], \operatorname{Cat}[\boldsymbol{D}_v; \boldsymbol{Q}_v]) \\
=&\operatorname{softmax}(\boldsymbol{h_{r}}\boldsymbol{W}_q \operatorname{Cat}[\boldsymbol{D}_k; \boldsymbol{Q}_k]^{\top})
\left[\begin{array}{c}
\boldsymbol{D}_v \\
\boldsymbol{Q}_v
\end{array}\right] \\
=&\frac{ \exp \left(\boldsymbol{h_{r}}\boldsymbol{W}_q \boldsymbol{Q}_k^{\top} \right) \boldsymbol{Q}_v + \exp \left(\boldsymbol{h_{r}}\boldsymbol{W}_q \boldsymbol{D}_k^{\top} \right)\boldsymbol{D}_v}
{\sum_i \exp \left(\boldsymbol{h_{r}}\boldsymbol{W}_q \boldsymbol{D}_k^{\top}\right)_i+\sum_j \exp \left(\boldsymbol{h_{r}}\boldsymbol{W}_q \boldsymbol{Q}_k^{\top} \right)_j} \\
=&\frac{\sum_j \exp \left(\boldsymbol{h_{r}}\boldsymbol{W}_q \boldsymbol{Q}_k^{\top} \right)_j}{\sum_i \exp \left(\boldsymbol{h_{r}}\boldsymbol{W}_q \boldsymbol{D}_k^{\top}\right)_i+\sum_j \exp \left(\boldsymbol{h_{r}}\boldsymbol{W}_q \boldsymbol{Q}_k^{\top} \right)_j} \\
&\times \frac{\exp \left(\boldsymbol{h_{r}}\boldsymbol{W}_q \boldsymbol{Q}_k^{\top} \right)}{\sum_j \exp \left(\boldsymbol{h_{r}}\boldsymbol{W}_q \boldsymbol{Q}_k^{\top} \right)_j} \boldsymbol{Q}_v \\
&+\frac{\sum_i \exp \left(\boldsymbol{h_{r}}\boldsymbol{W}_q \boldsymbol{D}_k^{\top}\right)_i}{\sum_i \exp \left(\boldsymbol{h_{r}}\boldsymbol{W}_q \boldsymbol{D}_k^{\top}\right)_i+\sum_j \exp \left(\boldsymbol{h_{r}}\boldsymbol{W}_q \boldsymbol{Q}_k^{\top} \right)_j} \\
&\times \frac{\exp \left(\boldsymbol{h_{r}}\boldsymbol{W}_q \boldsymbol{D}_k^{\top} \right)}{\sum_i \exp \left(\boldsymbol{h_{r}}\boldsymbol{W}_q \boldsymbol{D}_k^{\top}\right)_i} \boldsymbol{D}_v\\
=&\frac{\sum_j \exp \left(\boldsymbol{h_{r}}\boldsymbol{W}_q \boldsymbol{Q}_k^{\top} \right)_j}{\sum_i \exp \left(\boldsymbol{h_{r}}\boldsymbol{W}_q \boldsymbol{D}_k^{\top}\right)_i+\sum_j \exp \left(\boldsymbol{h_{r}}\boldsymbol{W}_q \boldsymbol{Q}_k^{\top} \right)_j} \\
&\times \operatorname{softmax}(\boldsymbol{h_{r}}\boldsymbol{W}_q \boldsymbol{Q}_k^{\top}) \boldsymbol{Q}_v\\
&+\frac{\sum_i \exp \left(\boldsymbol{h_{r}}\boldsymbol{W}_q \boldsymbol{D}_k^{\top}\right)_i}{\sum_i \exp \left(\boldsymbol{h_{r}}\boldsymbol{W}_q \boldsymbol{D}_k^{\top}\right)_i+\sum_j \exp \left(\boldsymbol{h_{r}}\boldsymbol{W}_q \boldsymbol{Q}_k^{\top} \right)_j} \\
&\times \operatorname{softmax}(\boldsymbol{h_{r}}\boldsymbol{W}_q \boldsymbol{D}_k^{\top}) \boldsymbol{D}_v \\
=&(1-\lambda(\boldsymbol{h_{r}})) \operatorname{softmax}(\boldsymbol{h_{r}}\boldsymbol{W}_q \boldsymbol{Q}_k^{\top}) \boldsymbol{Q}_v\\
&+\lambda(\boldsymbol{h_{r}}) \operatorname{softmax}(\boldsymbol{h_{r}}\boldsymbol{W}_q \boldsymbol{D}_k^{\top}) \boldsymbol{D}_v \\
=&(1-\lambda(\boldsymbol{h_{r}})) \underbrace{\operatorname{Att}\left(\boldsymbol{h_{r}}\boldsymbol{W}_q, \boldsymbol{Q}_k, \boldsymbol{Q}_v\right)}_{\text {outcome from }\boldsymbol{q}}\\
&+\lambda(\boldsymbol{h_{r}}) \underbrace{\operatorname{Att}\left(\boldsymbol{h_{r}}\boldsymbol{W}_q, \boldsymbol{D}_k, \boldsymbol{D}_v\right)}_{\text {outcome from } \boldsymbol{demos}},
\end{aligned}
\label{equation: ICL_new_detail}
\end{equation}
where:
\begin{equation}
\small
\lambda(\boldsymbol{h_{r}})=\frac{\sum_i \exp \left(\boldsymbol{h_{r}}\boldsymbol{W}_q \boldsymbol{D}_k^{\top}\right)_i}{\sum_i \exp \left(\boldsymbol{h_{r}}\boldsymbol{W}_q \boldsymbol{D}_k^{\top}\right)_i+\sum_j \exp \left(\boldsymbol{h_{r}}\boldsymbol{W}_q \boldsymbol{Q}_k^{\top} \right)_j}
\label{equation: ICL_lambda_detail}
\end{equation}

\section{Inverse-scaling Phenomena with Gemini}
\begin{figure*}[!htb]
    \centering
    \subfigure[Algebra]{\includegraphics[width=0.49\hsize, height=0.25\hsize]{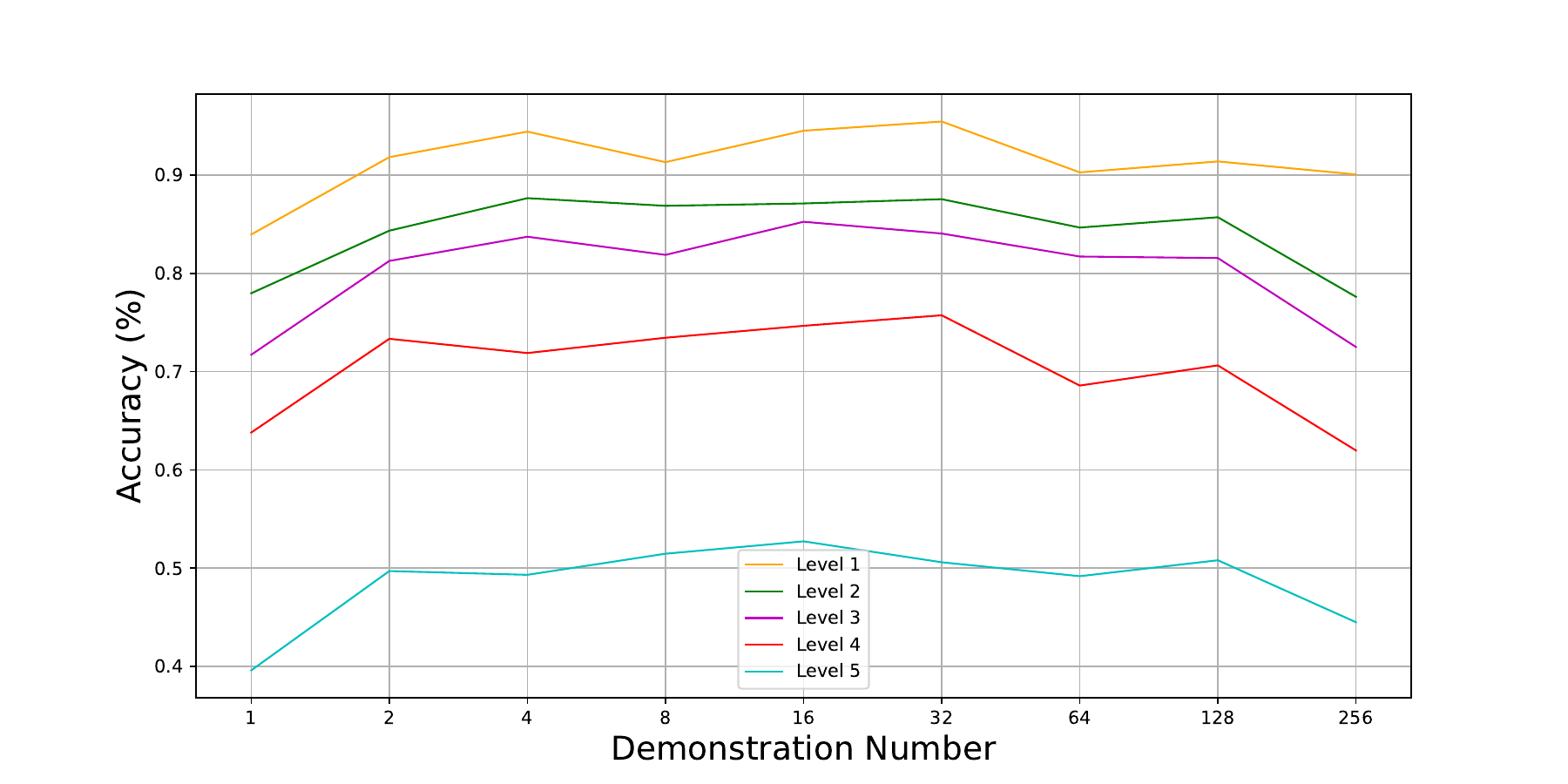}}
    \subfigure[Prealgebra]{\includegraphics[width=0.49\hsize, height=0.25\hsize]{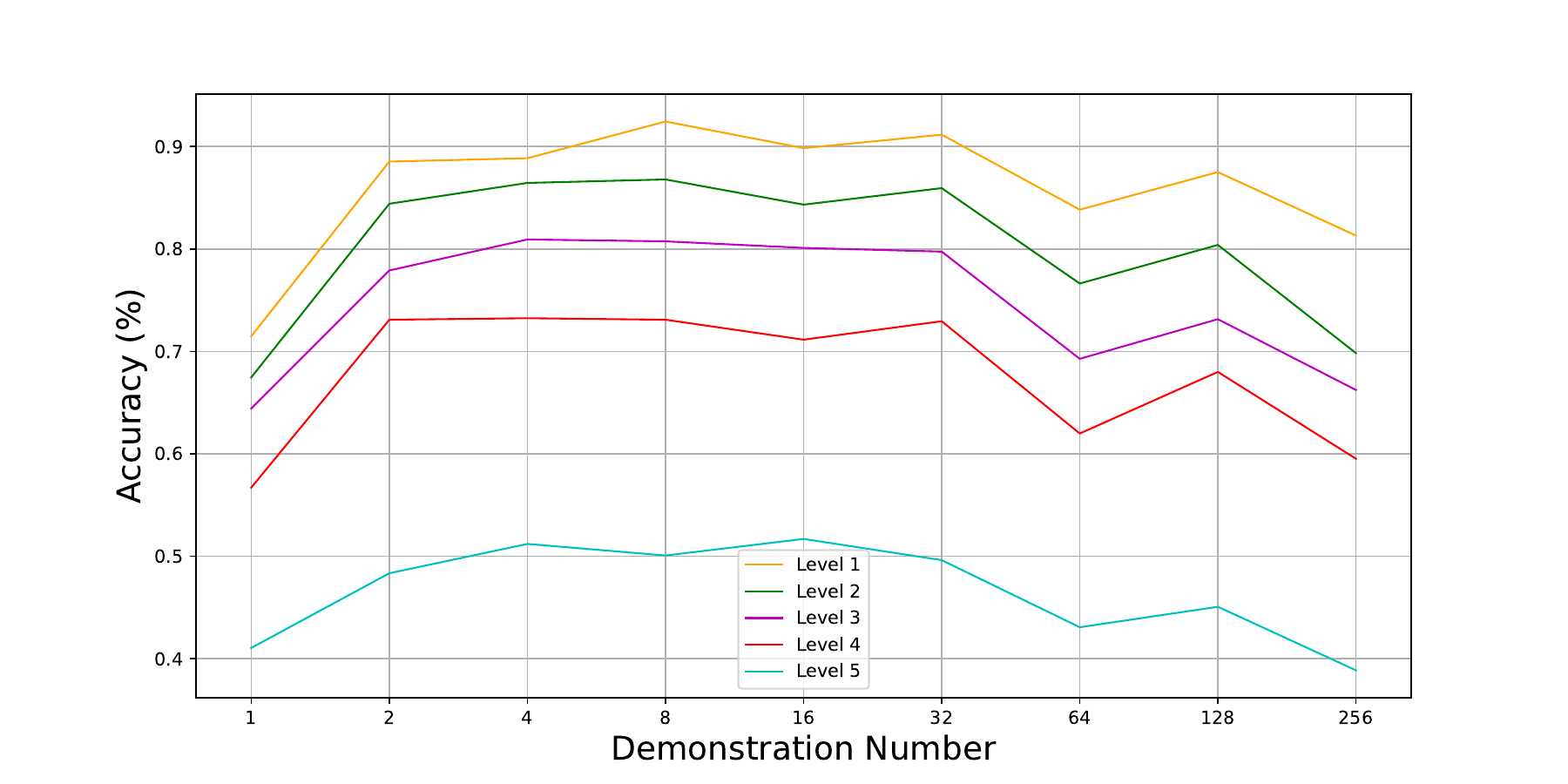}}
    \subfigure[Counting and Probability]{\includegraphics[width=0.49\hsize, height=0.25\hsize]{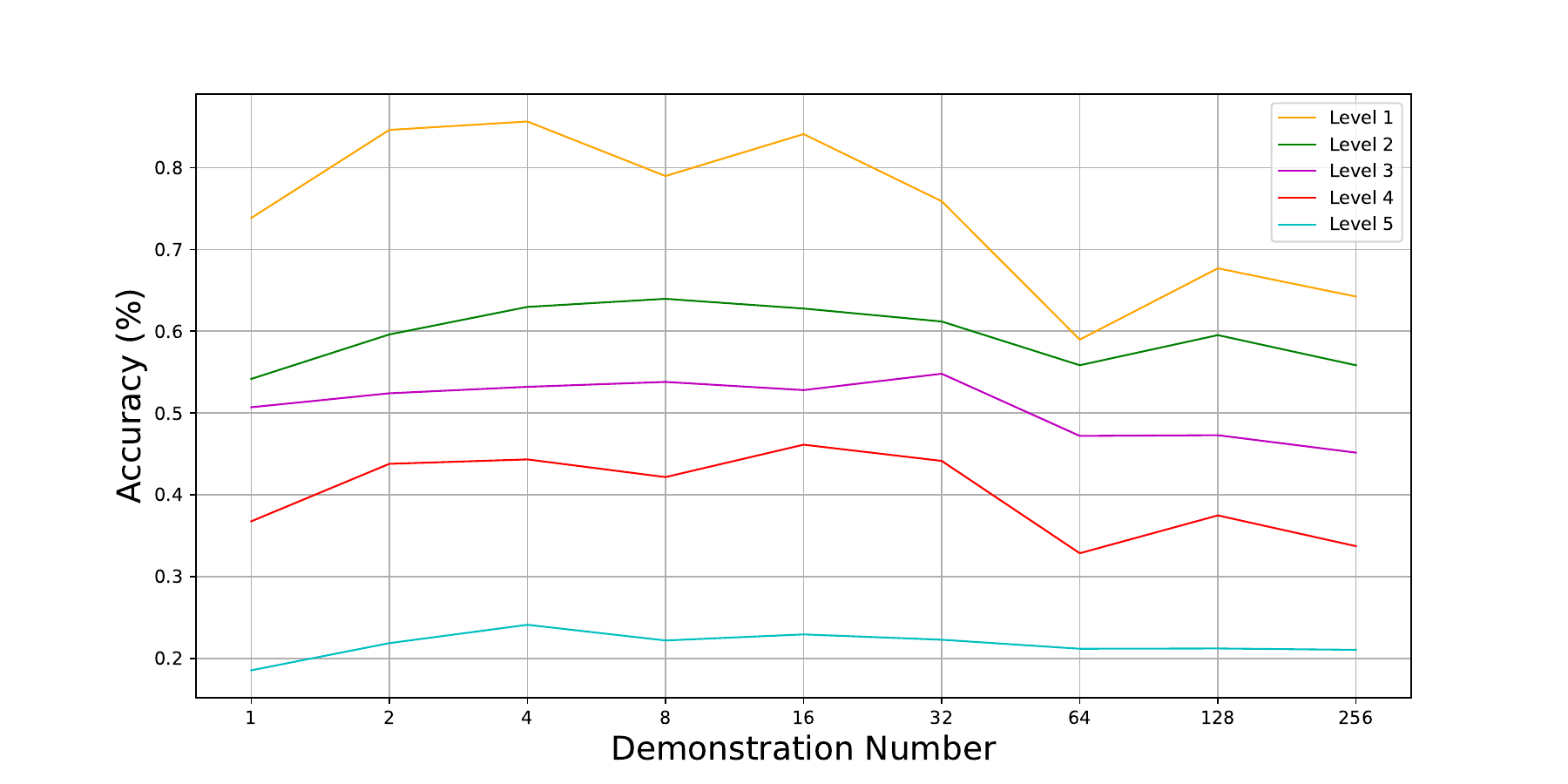}}
    \subfigure[Geometry]{\includegraphics[width=0.49\hsize, height=0.25\hsize]{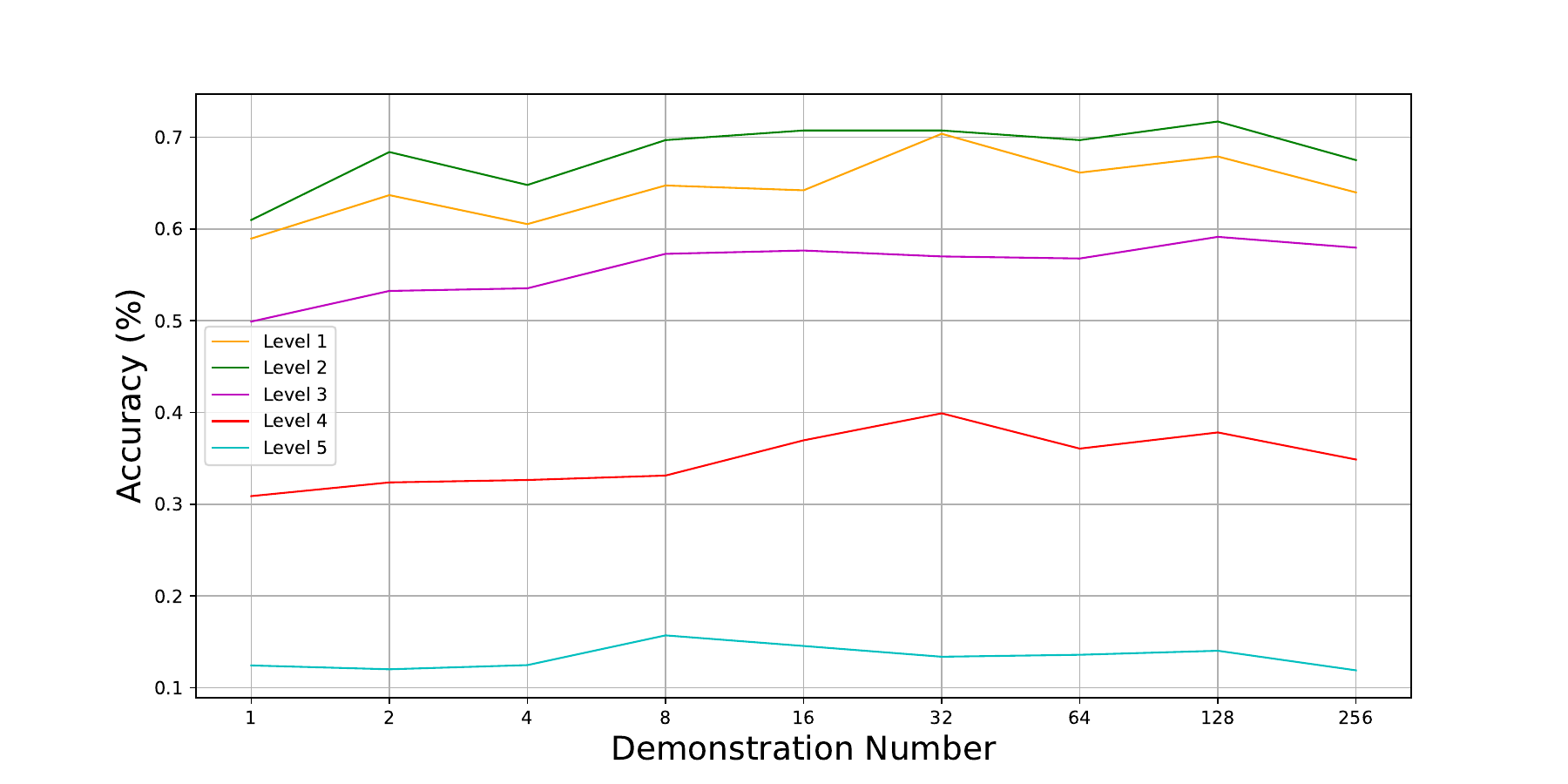}}
    \subfigure[Intermediate Algebra]{\includegraphics[width=0.49\hsize, height=0.25\hsize]{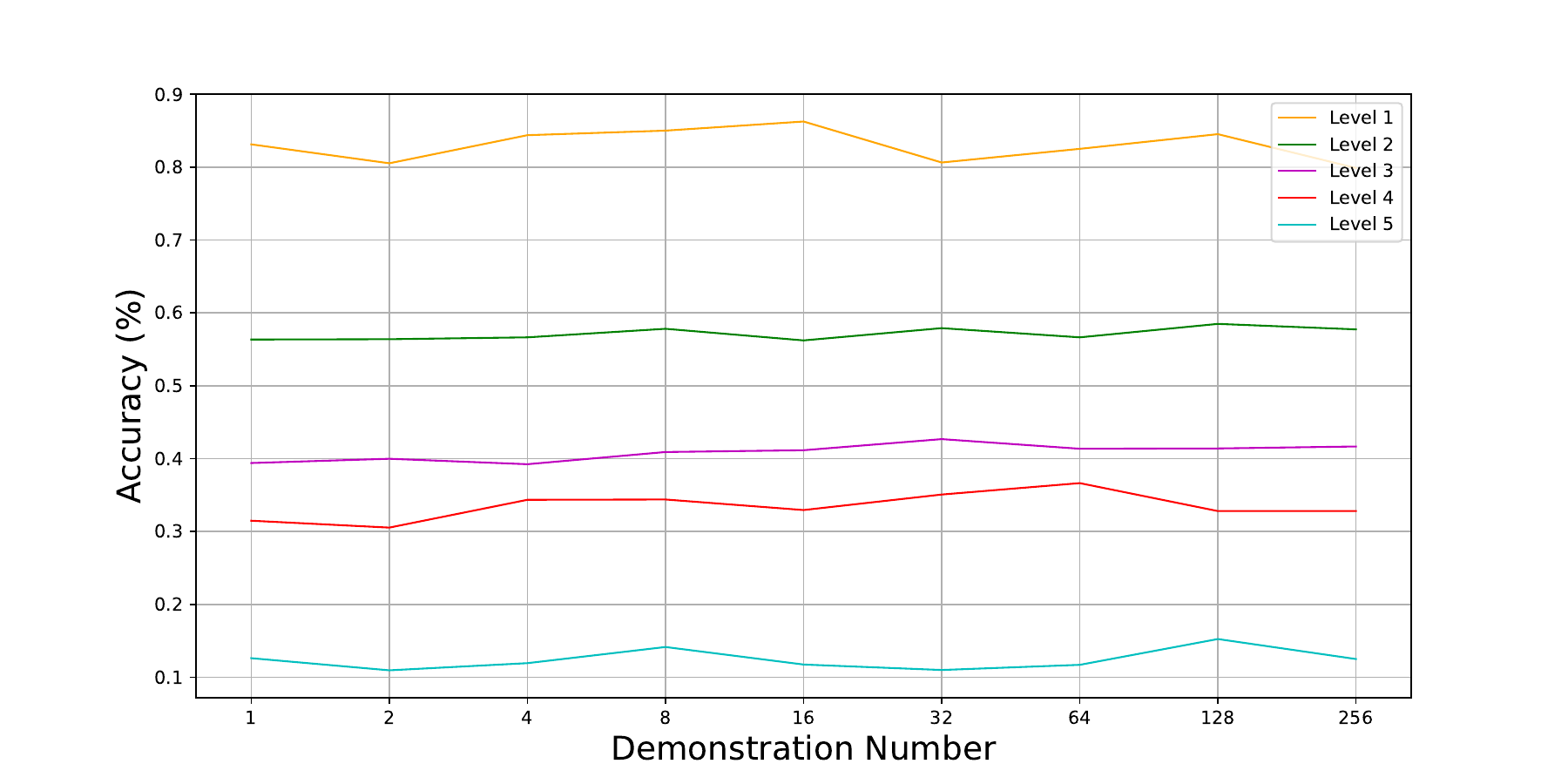}}
    \subfigure[Number Theory]{\includegraphics[width=0.49\hsize, height=0.25\hsize]{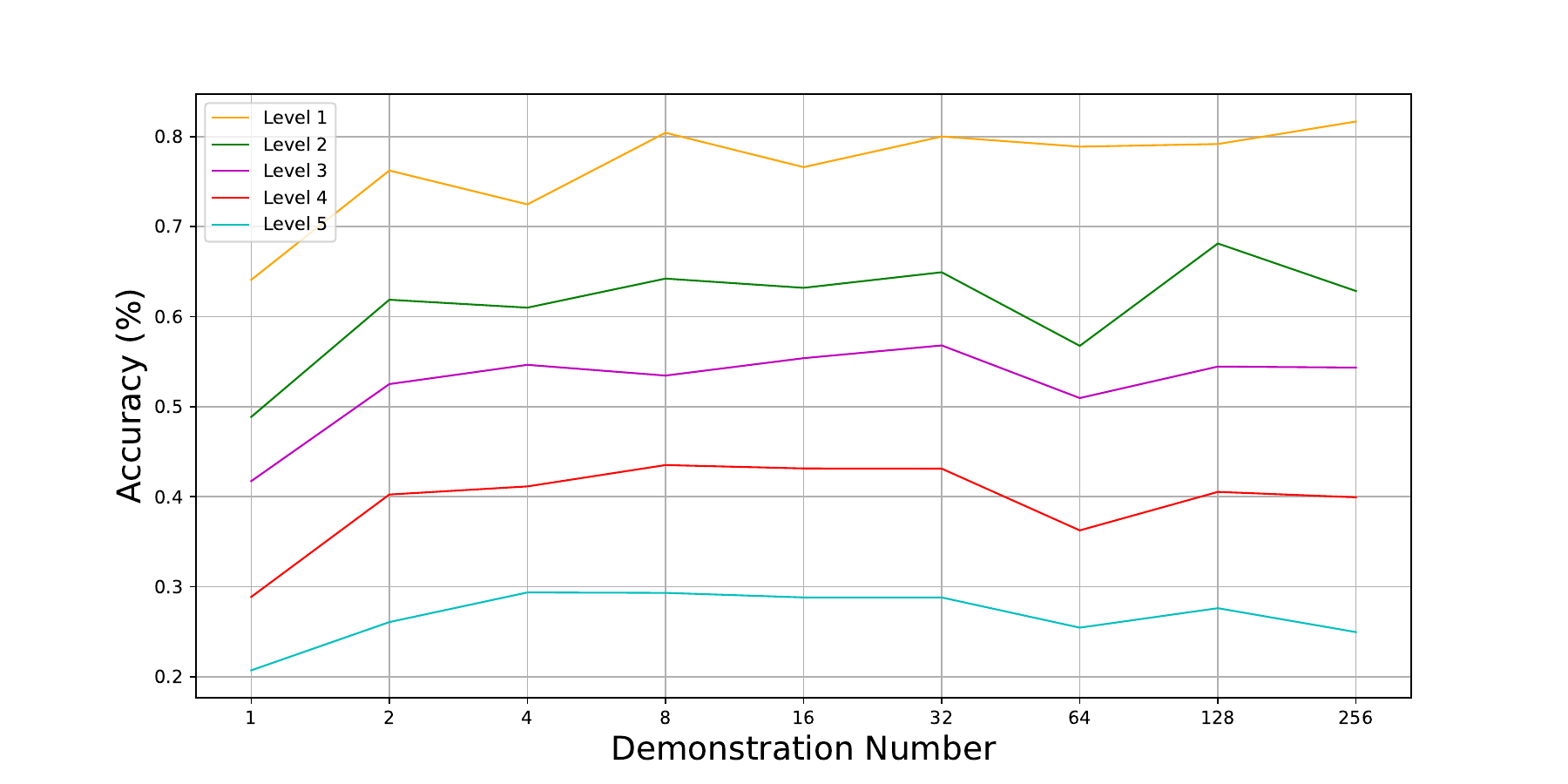}}
    \subfigure[Precalculus]{\includegraphics[width=0.49\hsize, height=0.25\hsize]{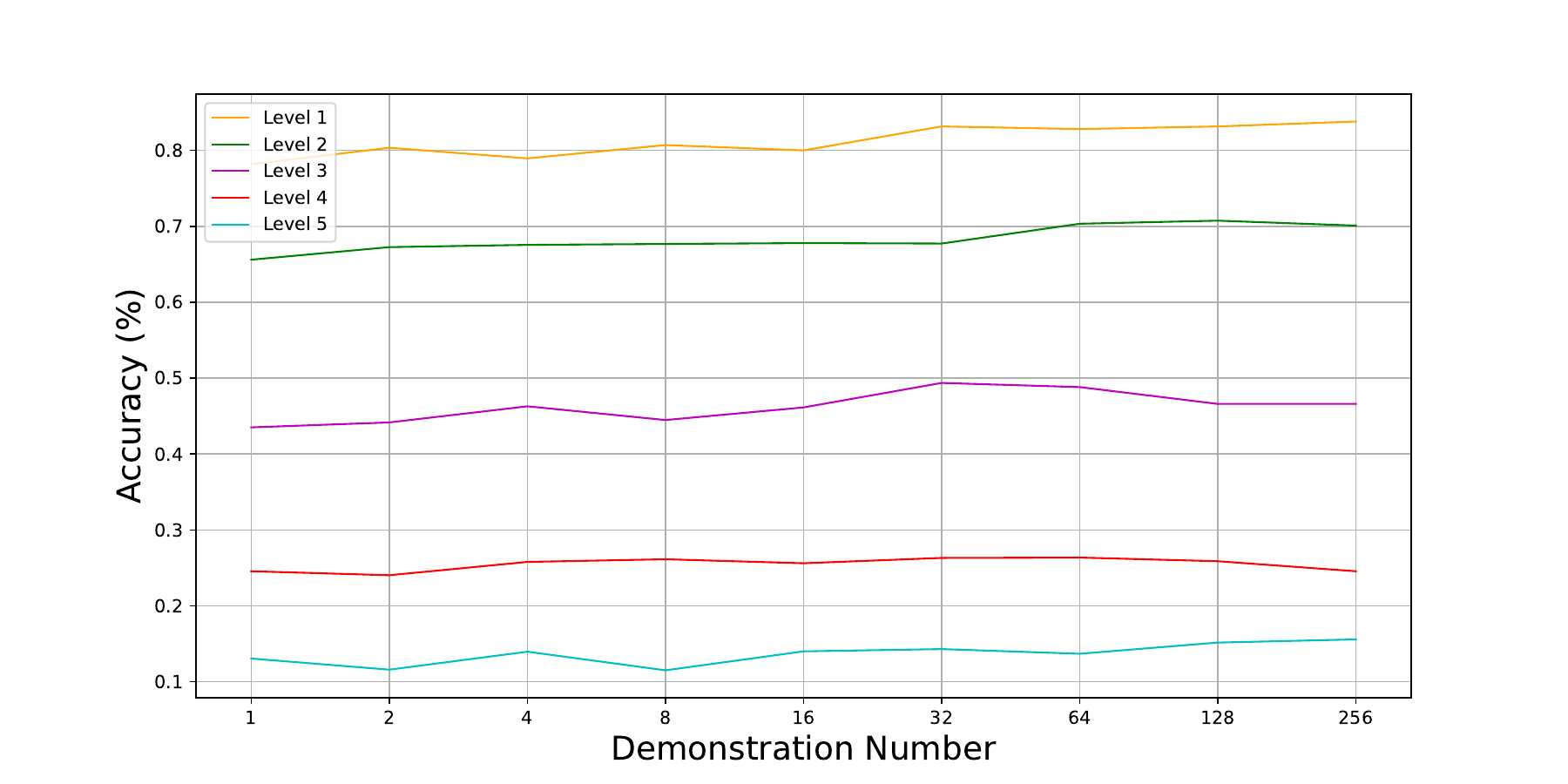}}
    \subfigure[Average]{\includegraphics[width=0.49\hsize, height=0.25\hsize]{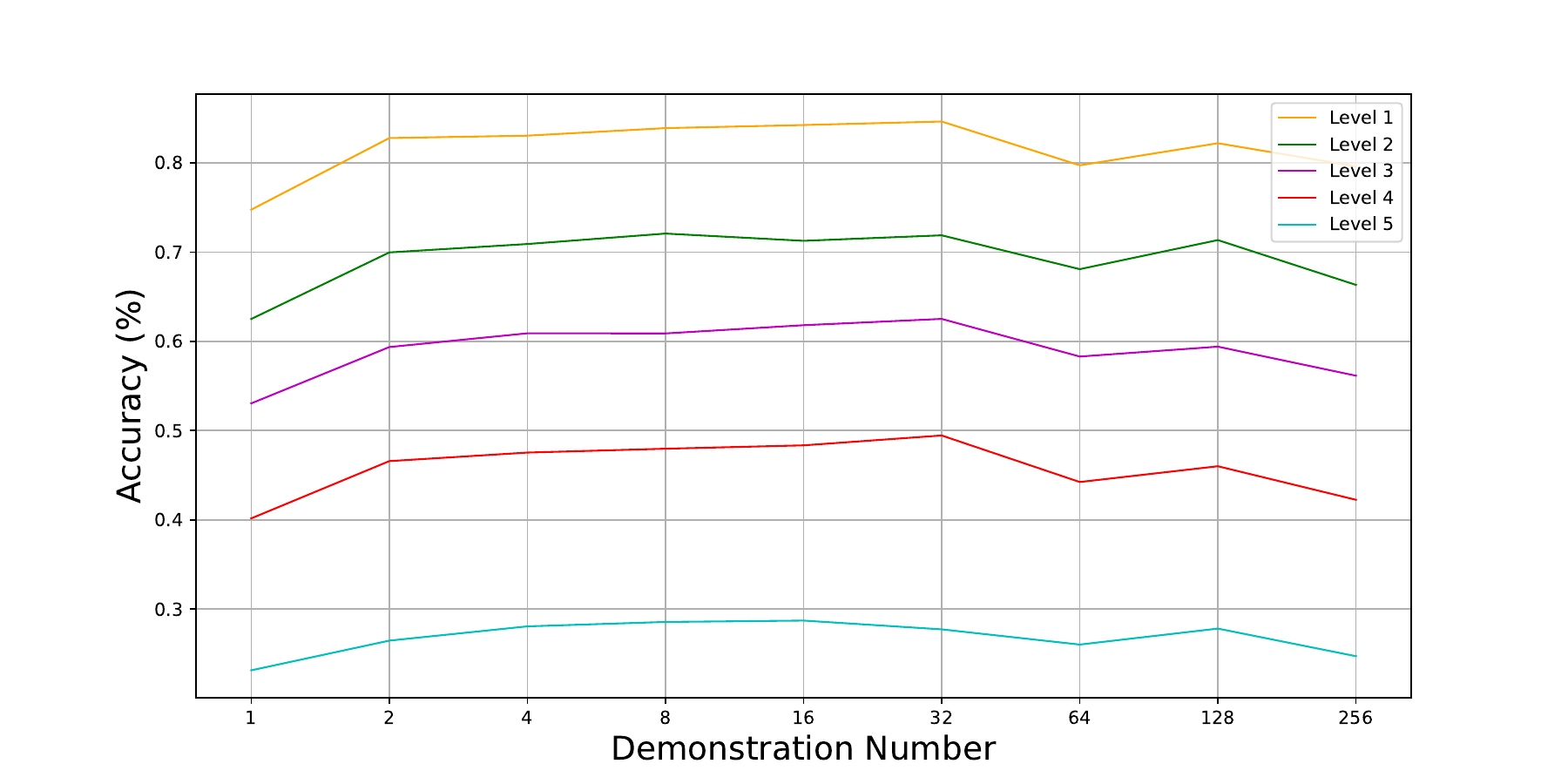}}\label{appendixfig:avg}
    \setlength{\belowcaptionskip}{0pt}
    \caption{Performance of Gemini on different subset of MATH with varying demonstration numbers.}
    \label{appendixfig: gemini}
\end{figure*}

Due to the limitations of computational resources and the unavailability of closed-source models, our experiments are primarily conducted on 7-8B open source LLMs. However, by utilizing APIs, we additionally explore the performance changes of more powerful models as the number of demonstrations increased, further validating the generalizability of the argument that LLMs are not stable many-shot learners.
We choose to experiment with \textsc{Gemini 1.5 Pro} for its long available context window (1M tokens). We test \textsc{Gemini 1.5 Pro} on MATH benchmark \citep{MATH}, which contains 7 subsets with 5 difficulty levels that can thoroughly evaluating the math reasoning abilities of LLMs. We use greedy searching decoding strategy with and report the outcomes averaged over 5 runs for credible results. As shown in Figure~\ref{appendixfig: gemini}, obvious inverse-scaling phenomenon appears in 5 out of 7 subsets, with Precalculus and Intermediate Algebra as exceptions. This validates the generalizability of the argument that LLMs are not stable many-shot learners. Meanwhile, we observe that across different difficulty levels, \textsc{Gemini 1.5 Pro} presents similar performance changing trends. Figure~\ref{appendixfig: gemini} clearly shows such phenomenon. This indicates that the task difficulty does not affects the optimal demonstration number of certain task. 

\begin{table}[t]
    \renewcommand\arraystretch{1.3}
    \small
    \centering
    \setlength{\tabcolsep}{0.13em} 
    \begin{tabular}{cccccc}
    \toprule
    \textbf{Filtering Threshold}&0.0 &0.1 &0.2  & 0.3&0.4\\
    \midrule
     \textsc{FocusICL}&11.90&12.28&12.03&12.05&11.88\\
    \bottomrule
    \end{tabular}
    \caption{Accuracy (\%) of \textsc{longchat-7b-v1.5-32k} when applying \textsc{FocusICL} with varying filtering threshold and batch size as 8.}
    \label{appendixtab: Main results1}
\end{table}
\begin{table}[t]
    \renewcommand\arraystretch{1.3}
    \small
    \centering
    \setlength{\tabcolsep}{0.13em} 
    \begin{tabular}{cccccc}
    \toprule
    \textbf{Batch Size}&2 &4 &8  & 16&80\\
    \midrule
     \textsc{FocusICL}&10.46&10.99&12.28&11.45&11.00\\
    \bottomrule
    \end{tabular}
    \caption{Accuracy (\%) of \textsc{longchat-7b-v1.5-32k} when applying \textsc{FocusICL} with varying batch sizes and filtering threshold as 0.1. It should be noted that the overall demonstration number is fixed as 80.}
    \label{appendixtab: Main results2}
\end{table}

\begin{table}[ht]
    \renewcommand\arraystretch{1.3}
    \small
    \centering
    \setlength{\tabcolsep}{0.13em} 
    \begin{tabular}{cccccc}
    \toprule
    \textbf{Method}&\textbf{CSQA} &\textbf{PIQA} &\textbf{CountA}  & \textbf{ARC}&\textbf{GSM8K}\\
    \midrule
     $N$&128&96&448&108&80\\
    \bottomrule
    \end{tabular}
    \caption{The total demonstration number $N$ of different benchmarks in our experiments.}
    \label{appendixtab: demonum}
\end{table}

\begin{table}[ht]
    \renewcommand\arraystretch{1.3}
    \small
    \centering
    \setlength{\tabcolsep}{0.13em} 
    \begin{tabular}{lccccc}
    \toprule
    \textbf{Method}&\textbf{CSQA} &\textbf{PIQA} &\textbf{CountA}  & \textbf{ARC}&\textbf{GSM8K}\\
    \midrule
     Training size&9741&16113&3000&2241&7473\\
     Testing size&1221&1838&1000&567&1319\\
    \bottomrule
    \end{tabular}
    \caption{Benchmark Statistics.}
    \label{appendixtab: data_statistics}
\end{table}

\begin{table}[t]
    \renewcommand\arraystretch{1.3}
    \small
    \centering
    \setlength{\tabcolsep}{0.13em} 
    \begin{tabular}{lcccccc}
    \toprule
    \multirow{2}{*}{Model}&\multicolumn{2}{c}{LONGCHAT-7B} &\multicolumn{2}{c}{VICUNA-7B}&\multicolumn{2}{c}{LLAMA-3-8B}\\
    &\multicolumn{2}{c}{-V1.5-32K}&\multicolumn{2}{c}{-V1.5-16K}&\multicolumn{2}{c}{-INSTRUCT}\\
    \cdashline{1-7}
    Params&$p$&$B$&$p$&$B$&$p$&$B$\\
    \midrule
     CSQA&0.1&32&0.2&16&0.4&32\\
     PIQA&0.1&32&0.1&8&0.4&2\\
     CountA&0.4&112&0.4&224&0.4&112\\
     ARC&0.4&16&0.4&0.1&0.4&12\\
     GSM8K&0.1&8&0.1&8&0.4&8\\
    \bottomrule
    \end{tabular}
    \caption{The results of hyperparameter searching strategy across varing tasks and LLMs.}
    \label{appendixtab: hyper}
\end{table}

\section{Further Discussions}
FocusICL can be seen as a method that achieves performance gains through increased computation (more demonstrations). Similar approaches include Self-Consistency \citep{SC0,SC3,SC2,SC1} and Chain-of-Thought \citep{CoT}. In our experiments, we have confirmed that the gains brought by \textsc{FocusICL} are decoupled from those of Chain-of-Thought. We will further explore the interplay between \textsc{FocusICL} and other methods in the future.

We tested the performance of \textsc{FocusICL} in tasks such as QA and inference in the experimental section. In the future, we will delve into exploring the application of \textsc{FocusICL} in evaluation \citep{eva1,eva2,eva3} and dialogue \citep{dia1,dia2,dia3} tasks.

\section{Prompt Template}
\label{prompt}
The following is a template ICL input format when demonstration number is 2.
\begin{quote}
{\itshape
\#\#\# Human: I'm getting warm because I increased the thermostat in my bedroom.  What might I be doing soon? Answer Choices: (a) feeling comfortable (b) overheat (c) increase of temperature (d) pleasure (e) starting fire
\\
\\
\#\#\# Assistant: A
\\
\\
\#\#\# Human: Where might I hear and see information on current events? Answer Choices: (a) internet (b) television (c) newspaper (d) book (e) radio
\\
\\
\#\#\# Assistant: B
\\
\\
\#\#\# Human: If somebody buys something and gives it to me as a free gift, what is the cost status of the gift? Answer Choices: (a) deadly (b) imprisoned (c) paid for (d) expensive (e) in prison
\\
\\
\#\#\# Assistant:
}
\end{quote}

\end{document}